\documentclass[10pt,twocolumn,letterpaper]{article}

\usepackage[final,algorithms]{wacv}      % To produce the REVIEW version for the algorithms track
\usepackage{graphicx}
\usepackage{amsmath}
\usepackage{amssymb}
\usepackage{booktabs}

\usepackage[pagebackref,breaklinks,colorlinks]{hyperref}

\usepackage{multirow}
\usepackage{hyperref}
\usepackage{autobreak}
\usepackage{amssymb}% http://ctan.org/pkg/amssymb
\usepackage{pifont}% 

\usepackage[ruled,linesnumbered]{algorithm2e} %algorithm
\usepackage{textcomp}

\usepackage{float}

\usepackage{amsmath,amssymb,amsfonts,dsfont,bm,bbm,mathrsfs,pifont}
\usepackage{xcolor}
\usepackage{multirow}
\usepackage{hyperref}
\usepackage{autobreak}
\usepackage{booktabs}
\usepackage{enumitem}

\usepackage{amssymb}% http://ctan.org/pkg/amssymb
\usepackage{pifont}% http://ctan.org/pkg/pifont
\newcommand{\cmark}{\ding{51}}%

\renewcommand{\paragraph}[1]{\vspace{0.5pt}\par\noindent\textbf{#1}~}

\def\figureautorefname{Fig.}

\usepackage[capitalize]{cleveref}
\crefname{section}{Sec.}{Secs.}
\Crefname{section}{Section}{Sections}
\Crefname{table}{Table}{Tables}
\crefname{table}{Tab.}{Tabs.}

\def\wacvPaperID{2598}
\def\confName{WACV}
\def\confYear{2025}

\begin{document}

\abovedisplayskip=2pt
\belowdisplayskip=3pt
\abovedisplayshortskip=1pt
\belowdisplayshortskip=2pt

\setlength{\floatsep}{8pt}
\setlength{\textfloatsep}{8pt}
\setlength{\dblfloatsep}{8pt}
\setlength{\dbltextfloatsep}{8pt}

% \title{Registration Guided Cross Teaching and Contrastive learning for Semi-supervised Medical Image Segmentation}
\title{Learning Semi-Supervised Medical Image Segmentation\\from Spatial Registration}

\author{Qianying Liu\\
University of Glasgow\\
{\tt\small 2665227L@student.gla.ac.uk}
\and
Paul Henderson$^\dagger$\\
University of Glasgow\\
{\tt\small paul.henderson@glasgow.ac.uk}
\and
Xiao Gu \\
University of Oxford\\ 
{\tt\small xiao.gu17@imperial.ac.uk}
\and
Hang Dai \\
University of Glasgow\\
{\tt\small hang.dai@glasgow.ac.uk}
\and
Fani Deligianni$^\dagger$ \\
University of Glasgow\\
{\tt\small fani.deligianni@glasgow.ac.uk}
}

\maketitle

%%%%%%%%% ABSTRACT
\begin{abstract}

Semi-supervised medical image segmentation has shown promise in training models with limited labeled data and abundant unlabeled data. However, state-of-the-art methods ignore a potentially valuable source of unsupervised semantic information---spatial registration transforms between image volumes. To address this, we propose CCT-R, a contrastive cross-teaching framework incorporating registration information. To leverage the semantic information available in registrations between volume pairs, CCT-R incorporates two proposed modules: Registration Supervision Loss (RSL) and Registration-Enhanced Positive Sampling (REPS). The RSL leverages segmentation knowledge derived from transforms between labeled and unlabeled volume pairs, providing an additional source of pseudo-labels. REPS enhances contrastive learning by identifying anatomically-corresponding positives across volumes using registration transforms. Experimental results on two challenging medical segmentation benchmarks demonstrate the effectiveness and superiority of CCT-R across various semi-supervised settings, with as few as one labeled case.
Our code is available at \href{https://github.com/kathyliu579/ContrastiveCross-teachingWithRegistration}{\textcolor[RGB]{240, 80,168}{https://github.com/kathyliu579/ContrastiveCross-teachingWithRegistration}}.
\end{abstract}
\def\thefootnote{$\dagger$}\footnotetext{Equal advising}
\def\thefootnote{}\footnotetext{This work was supported in part by the China Scholarship Council and EPSRC (EP/W01212X/1).}
\section{Introduction}

Semantic segmentation is a foundational task in medical image analysis.
However, supervised methods require meticulously annotated images, which are expensive and time-consuming to obtain. Alternatively, \textit{Semi-Supervised Semantic Segmentation} (S4) minimizes the need for manual annotation by leveraging a large pool of unlabeled images alongside a limited set of labeled images \cite{10189101}. %It has emerged as a important approach in medical image analysis since it offers a compelling balance between segmentation accuracy and resource efficiency. 

Existing S4 methods try to extract useful information from unlabeled data in various ways. One line of work \cite{chen2020big,aralikatti2023dual} first performs self-supervised pretraining on unlabeled data to learn robust features, then fine-tunes with limited labeled data. 
Other works learn from unlabeled images via pseudo-labeling \cite{tarvainen2017mean,olsson2021classmix, french2019semi } or consistency regularization strategies \cite{fan2022ucc, ouali2020semi, peng2020deep}, both of which retrain the model using its own predictions on unlabeled images as pseudo-supervision.
Cross-teaching frameworks, like the teacher-student \cite{tarvainen2017mean} and student-student paradigms \cite{Chen2021a, Luo2022}, learn from unlabeled data by encouraging consistency of predictions between different network branches.
Supervised contrastive learning endows the S4 model with a stronger feature-extraction ability \cite{Hu2021,Wu2022Cross,zhao2023rcps,Chaitanya2023}, encouraging features of pixels with the same class (positives) to be similar, and features of different classes (negatives) to be dissimilar. 
State-of-the-art (SOTA) cross-teaching methods \cite{Liu_2023_BMVC} also incorporate pixel-wise contrastive learning on multi-scale feature maps. However, learning a robust representation from numerous unlabeled images remains challenging due to potential noise in pseudo-labels.

% to encourage the networks to have similar features.
% Recently, great success has been achieved by introducing pixel-wise contrastive learning to semantic segmentation, which endows the model with a stronger features-extracting ability. Specially, these methods project each pixel to representation space as a representation and regularize it in a fully supervised manner, i.e., aggregating the representations with the same class and separating them with different classes. 

% In this work, we extend the single-space supervision to a dual-space supervision for registration-based S4 called xxxx. 

\textit{Spatial registration} is a related task that aims to find dense spatial correspondences between pairs of 3D image volumes \cite{Maintz1998Survey,Balakrishnan2019}.
Many methods, both classical and learning-based, do not require manual supervision, but are based on comparing pixel intensities or features.
Still, spatial registration yields a wealth of semantic information, as points matched by the registration transformation should, in principle, have the same semantic labels.
Indeed, registration techniques are commonly used in brain image analysis to directly propagate a segmentation map from a template image to another \cite{logothetis2008we}.
Despite the wide use of spatial registration in medical image analysis, the potential of harnessing registration for S4 remains under-explored.

In this work, we investigate how to improve S4 by leveraging the rich semantic information inherently available through off-the-shelf spatial registration methods.
By integrating this information into contrastive cross-teaching frameworks \cite{Luo2022, Liu_2023_BMVC} which currently represent the SOTA in S4 for medical images, we propose a novel method \textit{CCT-R}, incorporating two techniques that give substantial improvements in S4 performance for medical images.
% We focus on improving cross-teaching methods and their contrastive variants, which currently represent the SOTA in S4 for medical images \cite{Luo2022, Liu_2023_BMVC}.
% We propose a novel CCT-R method to integrate registration information into contrastive cross-teaching framework, yielding substantial improvements in S4.

Firstly, we use registration-derived semantic information to generate additional pseudo-labels for unlabeled data, and introduce a new loss allowing these to guide the segmentation process.
This is beneficial since the accuracy of existing cross-teaching methods is limited by the quality of pseudo-labels predicted by each network and used to supervise the other; these pseudo-labels are typically very noisy during the early stages of training.
In contrast, registrations can be computed offline, prior to training, with relatively high accuracy.
We can then use registration transforms to transfer annotations from labeled to unlabeled volumes.
To mitigate poor-quality registrations, we develop a simple yet effective ‘best registration selection’ (BRS) strategy that uses cycle-consistency to identify the most useful registrations for generating high-quality labels, without requiring extra supervision. 
In this way, more reliable pseudo-labels are available early in the training process, which helps avoid confirmation bias from cross-teaching, accelerates learning, and improves final segmentation performance.

Secondly, we use registration to optimise the sampling of pairs during pixel-wise contrastive learning.
The SOTA contrastive cross-teaching S4 approach, MCSC \cite{Liu_2023_BMVC}, selects positive pairs based on (potentially noisy) pseudo-labels, and only within the current minibatch.
By employing registration transformations, we can go further, identifying spatially-corresponding pixels for each anchor point across different volumes.
This allows us to sample spatially positive pairs across volumes for contrast, even when their current pseudo-labels are incorrect, e.g.~early in training.
Furthermore, to increase the diversity of registration guided positives, and avoid the constraints imposed by batch size, we construct a memory-bank of feature maps from across multiple volumes.

In summary, our main contributions are as follows:
\begin{itemize}[itemsep=0.5pt,topsep=0.5pt, left=\parindent]
    \item We propose CCT-R, the first registration-guided method for semi-supervised medical image segmentation, by integrating registration with a contrastive cross-teaching framework.
    \item We introduce a novel registration supervision loss that enhances cross-teaching, by providing additional and informative registered pseudo-labels early in training, automatically selecting the best registered volumes. 
    % \item We show how registration can be used to efficiently sample anatomically-corresponding positive pairs for supervised contrastive learning. %, drawn from a memory-bank of feature maps.
    \item We show how registration can be used to mitigate the noisiness of pseudo labels in supervised contrastive learning, by adding anatomically-corresponding positive pairs regardless the currently predicted class.   
    % \item
    % A registration-enhanced positives sampling module is designed to select additional spatial positive pairs among a memory-bank of feature maps for supervised contrastive learning. 
    %\item We provide detailed experimental results on two S4 benchmarks (CT abdomen and cardiac MRI), measuring the benefits of different ways to incorporate registration information in the learning process, for different S4 models including UAMT \cite{yu2019uncertainty}, CPS \cite{Chen2021a}, CTS \cite{Luo2022}, and contrastive variants. 
\end{itemize}

Our evaluation demonstrates that each of these strategies enhances accuracy when combined with several recent S4 algorithms including UAMT \cite{yu2019uncertainty}, CPS \cite{Chen2021a}, CTS \cite{Luo2022}, and contrastive variants. Implementing both strategies simultaneously proves even more effective. Our proposed CCT-R (based on CTS) achieves SOTA performance across all settings with particularly impressive gains under minimal supervision conditions.  
With just a  single labeled case, CCT-R improves Dice coefficient (DSC) by 33.6$\%$ and reduces Hausdorff Distance (HD) by 32.8~mm on ACDC cardiac MRI segmentation \cite{bernard2018deep}, while on Synapse abdominal CT \cite{landman2015miccai} it improves DSC by 21.3$\%$ and HD by 58.1~mm.

\section{Related Work}

\paragraph{Consistency regularization in semi-supervised medical image segmentation.}
% \pmh{fewer words please!}
% Semi-Supervised learning for Medical Image Segmentation has the potential to significantly reduce the need for large amounts of labeled data by utilizing vast amounts of unlabeled data, guided by small number of labeled samples. However, closing the performance gap between supervised and semi-supervised approaches in medical image segmentation is challenging for several reasons. There is an empirical distribution mismatch between labeled and unlabeled data due to heterogeneity and imbalance in the data \cite{bai2023bidirectional}. This makes it difficult to leverage unlabeled data effectively to achieve prediction consistency. Additionally, relying on prediction models to generate pseudo image annotation pairs from unlabeled images often results in smooth label maps that do not accurately represent organ boundaries. \cite{liu2022reco}. Furthermore, techniques that are successful in semantic segmentation for natural images often fail in the medical imaging domain due to the requirement for pixel- and voxel-level accurate labeling \cite{Luo2022}. 
%Using prediction consistency as a regularizer and to extract information from unlabeled data is a mainstream strategy in S4.
% in addition to pseudo-labeling methods \cite{tarvainen2017mean,olsson2021classmix, french2019semi }. 
Semi-supervised learning is a very effective approach to address the challenge of limited annotations in medical image segmentation \cite{peng2020deep,bortsova2019semi,Luo2022,chen2020deep,lei2022semi}.
Researchers have proposed various consistency regularization approaches that enforce consistency between multiple branches, either through data augmentations \cite{bortsova2019semi, peng2020deep}, network architectures \cite{Luo2022}, or task configurations \cite{Wang2022}.
For instance, Bortsova~\etal{}~\cite{bortsova2019semi} encouraged consistency between the predicted masks and the input images under spatial transformations.
Peng \etal{} \cite{peng2020deep} used adversarial learning to encourage diverse predictions among a set of models, while Luo \etal{} \cite{Luo2022} leveraged Transformer-CNN consistency. 
However, most of these methods focus on prediction consistency for each single slice, overlooking feature relationships between different slices \cite{Liu_2023_BMVC}.
Additionally, relying on models to generate pseudo-labels often results in inaccurate organ boundaries \cite{liu2022reco}. 
Addressing these limitations remains an open challenge. 
Our CCT-R encourages both output and feature consistency between two branches \cite{Luo2022,Liu_2023_BMVC}, while uniquely using registration to provide richer information beyond cross-teaching alone.

\paragraph{Medical image registration.}
% \pmh{just enough background to give context -- learnt vs classical methods; what the inputs and outputs are}
Spatial registration is the process of aligning images from various sources, times, or patients to a common coordinate system \cite{Maintz1998Survey}, enabling tasks like automatic segmentation \cite{thor2011deformable, hautvast2006automatic}, mathematical modeling \cite{oh2014novel}, and functional imaging \cite{yamamoto2011impact}. 
Classical methods, such as those based on mutual information (MI) \cite{Viola1997}, and feature-based techniques like Demons registration \cite{Thirion1998}, align images by optimizing a cost function to minimize misalignment.
These approaches rely heavily on pixel intensities and anatomical features. 
Recent advances in deep learning have introduced learnt methods \cite{Balakrishnan2019,ding2023deep}, which automate feature extraction and optimization. These methods can be supervised (trained with labeled reference deformations) \cite{sokooti2017nonrigid, eppenhof2018pulmonary} or unsupervised (optimize similarity metrics without ground truth) \cite{Balakrishnan2019, hu2018weakly, ho2023unsupervised}. 
Both classical and learnt methods typically take a pair of images (fixed and moving) as input, and produce a transformation matrix or a dense deformation field that aligns them.
% Compared with classical approaches, learnt methods are faster and more flexible, making them vital in modern medical imaging.
% \pmh{there are also methods that learn a similarity descriptor, which are deep but slow...}
% \pmh{`flexible' is debateable -- for `one shot' learnt methods, they only work on the distribution of data they're trained on whereas elastix/itk are more general}

\paragraph{Combining segmentation and registration.}
% % multi tasks 
% A study \cite{andresen2022deep} introduces a novel CNN architecture, which is designed to simultaneously perform image registration and unsupervised segmentation of pathologies regions, enhancing the robustness and accuracy of registration in the presence of pathologies.
% \cite{liu2023co} the three networks collectively work to achieve registration and segmentation of pathological images under unsupervised conditions. 
% \cite{2023Ding} is designed to simultaneously perform multi-sequence cardiac magnetic resonance (MS-CMR) image registration and myocardial pathology segmentation. 
Segmentation and registration are closely related tasks that can complement each other, as both require extracting similar information from images.
Several methods achieve segmentation purely by propagating the labels from an atlas image to another after registration, such as for gray/white matter \cite{fischl2002whole} or V1/V2/IT \cite{benson2014correction} regions of brain, cardiac MR images \cite{lorenzo2002atlas} and liver CT \cite{rusko2009automatic}.
% Moreover, labeled atlas images can be used to provide segmentations of newly captured images, by registering those images to the atlas \pmh{cite -- some of the above}. 
Conversely, segmentation can provide additional supervision (beyond image intensities) for registration \cite{avants2009advanced}, as well as serve as a mean to evaluate registration results \cite{klein2005mindboggle}.
% multi tasks 
Consequently, many studies have explored joint training of deep networks for registration and segmentation across various supervision levels: unsupervised \cite{andresen2022deep,liu2023co}, fully supervised \cite{ 2023Ding, 9460972, beljaards2020cross,9615237}, few shot \cite{li2023prototypical, wang2024styleseg} and semi-supervised \cite{xu2019deepatlas}.
The most relevant to our CCT-R, DeepAtlas \cite{xu2019deepatlas}, jointly learns registration and S4 using 3D networks. However, they leverage neither established registration techniques nor modern S4 strategies like co-training and contrastive learning, limiting their approach to simpler anatomies (knee and brain).
% % treat singe task respectively 
% Additionally, a study \cite{li2023enabling} generated realistic ultrasound images to augment both segmentation and registration task (train alone) \pmh{so they synthesise images with a generative model; then use those images to train a segmentation model, and separately use to train a registration model? if it's really very separate like this, maybe don't mention this one }. 
%\pmh{the structure of this para is a bit messy: might be better to remove the introductory 2nd sentence (`for example...' since you then restate the same ideas here); but more this `moreover ... purely through...' sentence earlier}
Unlike these works, our approach does not aim to solve registration itself. Instead, it leverages an existing (imperfect) registration algorithms to boost the performance of S4.

\paragraph{Contrastive learning for segmentation and registration.}
Contrastive learning has been pivotal in self-supervised representation learning \cite{Chen2020,Grill2020,Tian2021,He2020}.
Early contrastive learning approaches focused on image-level (global) representations \cite{oord2018representation, he2020momentum,chen2020improved, grill2020bootstrap}, increasing similarity between positive pairs while differentiating negative pairs. 
To adapt contrastive learning to the segmentation task, which requires dense predictions, recent research has introduced pixel-level (local) self-supervised contrastive learning \cite{Xie2021, Wang2021}.
Some methods \cite{chaitanya2020contrastive} incorporate both local and global contrastive losses in segmentation.
% To mitigate the false negative predictions that are prone to occur in self-supervised local contrastive learning, 
These self-supervised methods are prone to false negative predictions \cite{khosla2020supervised}; to mitigate this, 
existing works \cite{Hu2021,Chaitanya2023,Liu_2023_BMVC} have explored supervised local contrastive learning. 
%In fact, it has also been shown that adding a supervised local contrastive loss at both global and local scales improves segmentation performance within a cross-teaching framework \cite{Liu_2023_BMVC}.
In the field of natural images, the integration of semi-supervised learning and contrastive learning has become a popular trend. This has lead to the development of one-stage, end-to-end models that eliminate the need for self-supervised pretraining \cite{Zhong2021,Yang2022,huang2023contrastive,10121455,long24eccv,10034936}. This approach has also been successfully applied to medical image segmentation \cite{Hu2021,Wu2022Cross,zhao2023rcps,Chaitanya2023, Basak_2023_CVPR}. 
Lastly, some works use self-supervised contrastive learning for registration, aiming to achieve high mutual information between fixed and moving images at the level of whole images \cite{liu2023contrastive} or patches \cite{song2022cross,dey2022contrareg}.
Unlike the above works, our CCT-R is the first to use registration information to guide contrastive sampling for S4.

\begin{figure*}[t]
\begin{center}
\vspace{-3pt}
\includegraphics[width=0.9\linewidth]{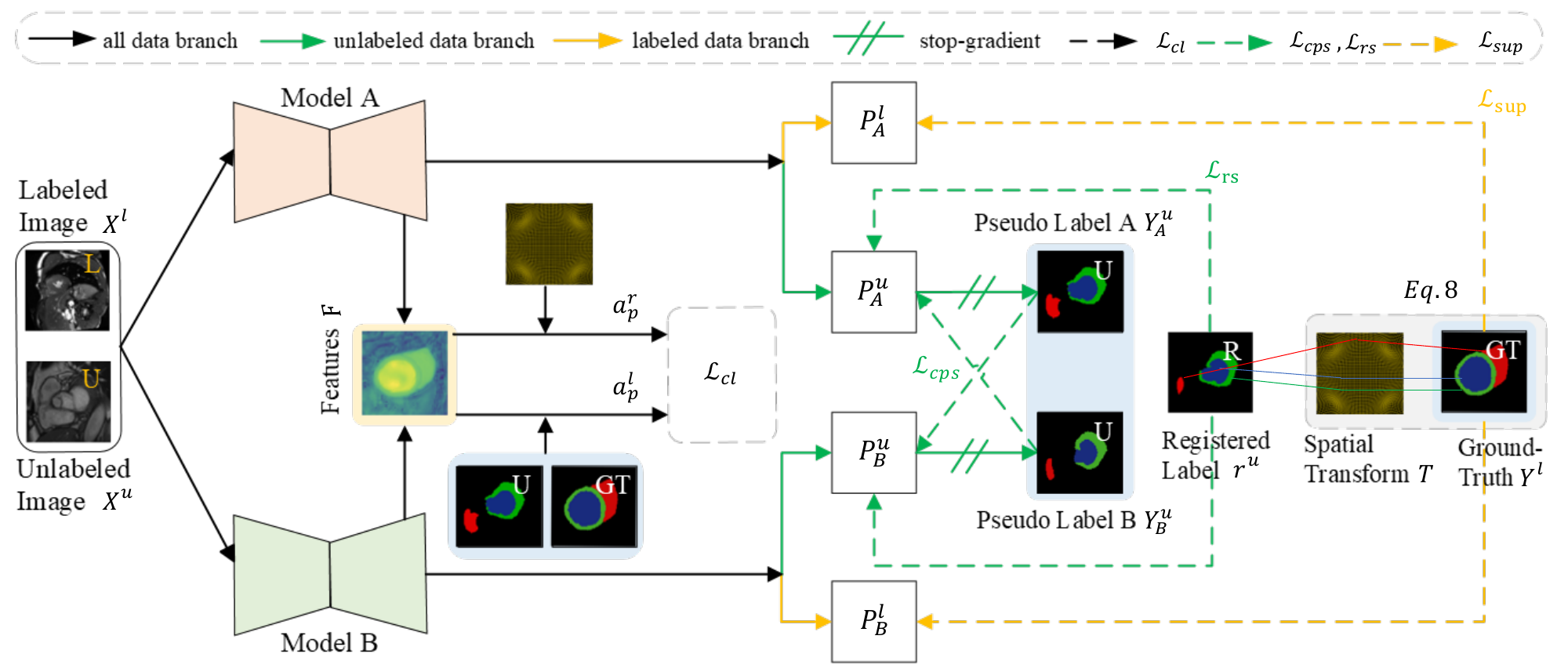}
    \end{center}
\vspace{-16pt}
\caption{The overall architecture of our framework for semi-supervised medical image segmentation.
}
% \vspace{-8pt}
\label{fig:framework}
\end{figure*}

\section{Method}

% In this section, we provide an overview of our registered contrast (RC)/cross teaching with registered contrast (CT-RC), followed by a detailed description of two main components: the Dual Spaces Supervised Contrast Module (DSSC) and the Registration Supervising Loss (RSL).
We first describe our problem setup and overall learning framework (Section~\ref{subsec:method-prelims}), which closely follows SOTA cross-teaching methods \cite{Chen2021a, Luo2022, Liu_2023_BMVC}. Next, we introduce the main technical contributions for our CCT-R: incorporating registration into the S4 framework (Section~\ref{subsec:method-reg-overview}), followed by a detailed description of how this is accomplished through a Registration Supervision Loss (RSL) (Section~\ref{subsec:method-reg-sup}) and by improving the quality of contrastive pairs with the Registration-Enhanced Positives Sampling (REPS) module 
(Section~\ref{subsec:method-reg-contrast}).

\subsection{Preliminaries}
\label{subsec:method-prelims}

% \pmh{only to describe the frameworks without registration, general enough to cover MT/CPS/CTS, contrastive is an optional extra bit -- then it is clear we are not describing any technical contribution here, just the learning setup (inspired by existing things) that we'll be improving on}

% S4 aims to obtain good segmentation performance by leveraging data comprising of few labeled 2D slices $X_{l} = \{(x_{i}^l,y_{i}^l)\}_{i=1}^K$ and many unlabeled slices $X_{u} = \{x_{j}^u\}_{j=1}^M$ (\ie $M \gg K$).

S4 aims to obtain good segmentation performance by leveraging data comprising of few labeled 2D slices \(D_l = \{(x_i^l, y_i^l)\}_{i=1}^K\) and many unlabeled slices \(D_u = \{x_j^u\}_{j=1}^M\) (\ie \(M \gg K\)). Let \(V = \{v_n\}_{n=1}^N\) represents the set of all 3D volumes, from which the set \(D = D_l \cup D_u\) is extracted. 
 % = \{x_i\}_{i=1}^{K+M}\
 
Our overall learning framework is similar to cross pseudo supervision \cite{Chen2021a, Luo2022} (\figureautorefname{}~\ref{fig:framework}), and the input is a minibatch $X = X^l \cup X^u$ including labeled images and unlabeled images. 
It uses two student models that are trained via a standard supervised loss $\mathcal{L}_{sup}$ on \(X^l\), and via a cross pseudo supervision loss $\mathcal{L}_{cps}$ on \(X^u\) where each network learns from the predictions of the other.

The supervised loss combines Dice and cross-entropy terms, similar to \cite{Liu_2023_BMVC, bai2023bidirectional}:
\begin{equation}
 \mathcal{L}_{sup} = - \frac1{K} \sum_{i=1} ^K \left(\mathcal{L}_{dice}(P^l_*, Y^l) + \mathcal{L}_{ce}(P^l_*, Y^l)\right).
\end{equation}
Here $P^l_*$ is the predicted class probability map of the labeled image batch $X^l$, calculated according to $P^l_* = C_*(E_*(X^l))$ where $E_*(\cdot)$ is a feature extractor, $C_*(\cdot)$ is a segmentation head yielding class probabilities for each pixel, $Y^l$ is the ground-truth label maps and $*$ denotes the model A or B.

The cross pseudo supervision loss $\mathcal{L}_{cps}$ \cite{Chen2021a} enables model A and model B teach each other on the unlabeled $X^{u}$, encouraging their respective predictions to be consistent. %, $i.e$, the model A's predictions guide the model B and vice-versa.
Specifically, we define
\begin{equation}
 \small
\label{CPS}
 \mathcal{L}_{cps (A)} = \mathcal{L}_{dice}( P^u_{A}, Y^u_{B}), \;\;\;
 \mathcal{L}_{cps (B)} = \mathcal{L}_{dice}( P^u_{B}, Y^u_{A}).
\end{equation}
Here the Dice loss $\mathcal{L}_{dice}$ for model A uses pseudo-labels $Y^u_{B}$ predicted by model B as its target, instead of ground-truth labels as in $\mathcal{L}_{sup}$.
Note that there is no gradient back-propagation between $P^u_{A}$ and $Y^u_{B}$ during training, nor between $P^u_{B}$ and $Y^u_{A}$.
In Section~\ref{subsec:method-reg-sup}, we will show how using spatial registration information can improve accuracy by providing additional pseudo-labels that are often less noisy than the cross teaching predictions.

\paragraph{Supervised contrastive learning.}
In addition, we optionally incorporate a supervised contrastive learning loss $\mathcal{L}_{cl}$, to better capture high-level semantic relationships between distant regions of different cases across the entire dataset.
Our contrastive loss follows \cite{Khosla2020}, but with the key difference that it contrasts pixel features instead of whole-image features.
We project each pixel to a shared embedding space then regularize in a supervised manner, encouraging features of anchor pixels to be similar to those of pixels having the same class (positives), and to be dissimilar to those of different classes (negatives).

Specifically, as shown in  \figureautorefname{}~\ref{fig:framework}, we extract a feature batch $F = F_{A} \cup F_{B}$, where $F_*=H_*(E_*(X))$ and $H_*(\cdot)$ is the projector. 
The choice of anchors, which serve as the comparison target of each class, has a great impact on learning; 
we therefore try to reduce the number of anchors with incorrect class labels.
For every class in the current mini-batch, we sample pixels with high top-1 probability value as anchors $A_c$ for class $c$, setting
\begin{equation}
A_c = \big\{ f_i \,|\, (y_i=c )\wedge (p_i>h) \big\},
\end{equation}
where $f_i$ is the $i$\textsuperscript{th} pixel feature in $F$, and the threshold $h$ for top-1 probability value is set to 0.5 to only exclude hard samples. 

The supervised contrastive loss $\mathcal{L}_{cl}$ is then computed as:
\begin{align}
\mathcal{L}_{cl} &= - \frac1{\lvert C\rvert} \sum_{c \in C} \frac1{\lvert \mathrm{an_c}\rvert} \sum_{a_{i} \in \mathrm{an_c}} \log\left\{ \frac {\exp(a_{i} \cdot a_p / \tau )}{\exp(a_{i} \cdot a_p / \tau ) + Z}\right\}
, \label{out} \\[-12pt]
Z &=  \sum_{\substack{j\in {C}\\ j\neq c}} \, \sum_{a_k \in {n_c^j}} \exp(a_{i} \cdot a_k / \tau ). \nonumber 

\end{align}
Here $C$ is the number of classes, $\mathrm{an_c} \subseteq A_c$ is the current anchor subset, \ie $N$ randomly sampled queries from the anchor set $A_c$, $a_i$ represents the $i^{th}$ anchor of class $c$, $n_c \subseteq N_c$ is the current negative set, \ie $O$ randomly sampled keys from $N_c$ (the negative set of class $c$), $n^j_c \in n_c$ is the subset of negative keys with class $j$, $j \neq c$, and $\tau$ is a temperature constant.
To prevent the background class from dominating the learning process, we limit the number of negative samples for each category. It ensures balanced contributions across classes and reduces memory usage, unlike \cite{Hu2021} which simply discards background features.
Note that in our experiments, $N=1000$ and $O=500$.
The positive key $a_p$ is given by calculating the average of all other pixels of the same class, \ie in the anchor set $A_c$:
\begin{equation}
% \small
 a_p^l = \frac1{\lvert A_c\rvert} \sum\limits_{a_i \in {A_c}} a_{i}. 
 \label{eq:supd-positives}
\end{equation}
Contrasting only an average positive instead of all positives is computationally cheaper, yet still allows reducing the average distance between the anchor and other samples of class $c$ \cite{liu2022reco}.
In Section~\ref{subsec:method-reg-contrast} we will show how using spatial registration information can provide additional positives for contrastive learning.

\begin{figure*}[t]
\begin{center}
\vspace{-2pt}
\includegraphics[width=0.85\linewidth]{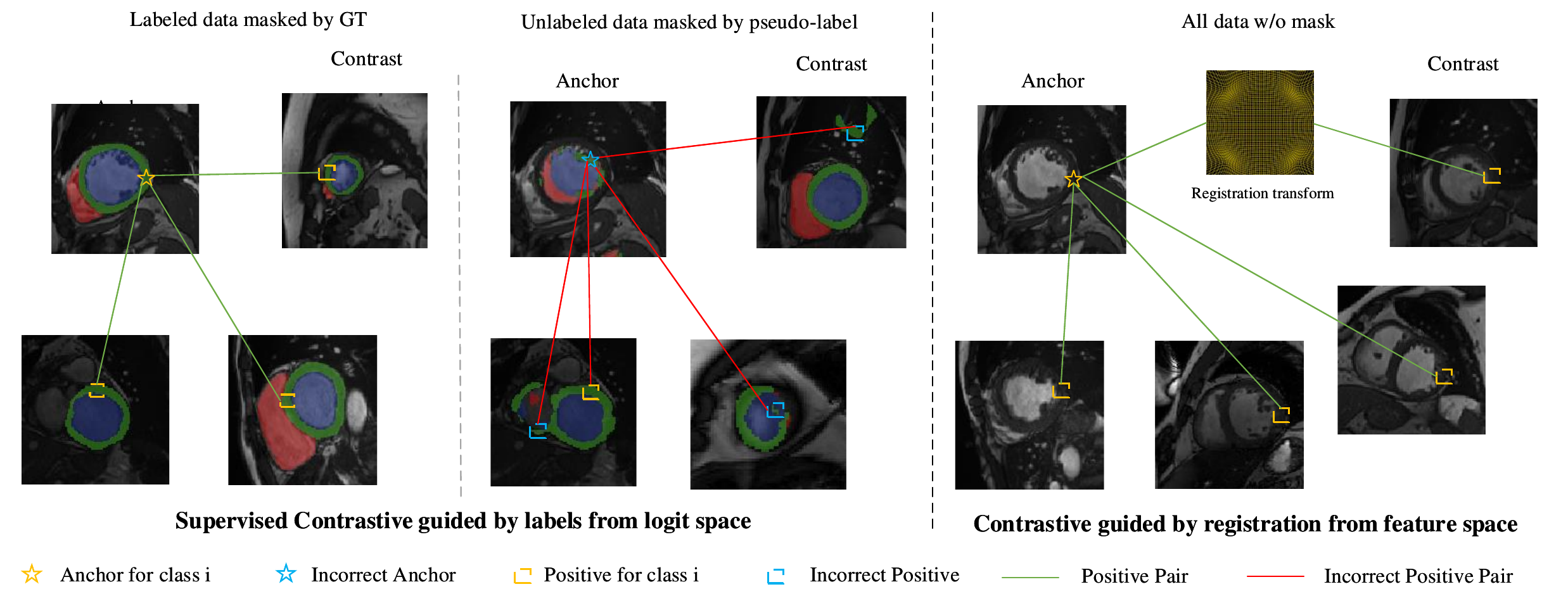}
    \end{center}
\vspace{-11pt}
\caption{\textbf{Supervised contrastive learning guided by labels \textit{vs.}~registration}: In the semi-supervised setting, for unlabeled data, the supervised contrastive loss uses pseudo-label information to select pairs. However, pseudo-labels are unreliable, especially early in training. For example, in the middle panel, the anchor is wrongly labeled as Myo (green), which leads to an incorrect learning signal, due to contrasting with positives correctly labeled as Myo. In contrast, registration finds the anatomically-closest point to the anchor in each 3D volume, without relying on label predictions from models, enabling the contrastive loss to perform correct comparisons between cases.
}
% \vspace{-2pt}
\label{fig:contrastive_learning}
\end{figure*}

\paragraph{Training and inference.}
The two models are trained simultaneously with separate losses.
The total training loss $\mathcal{L}_{A}$ for model A is:
\begin{equation}
 \mathcal{L}_{A} = \mathcal{L}_{sup(A)} + 
 w_{cps}\mathcal{L}_{cps (A)} + w_{cl}\mathcal{L}_{cl}.
\label{total_loss}
\end{equation}
and similarly for model B.
Here $w_*$ are weighting factors used to balance each loss term.
Overall, this setup yields comparable performance to the SOTA contrastive cross-teaching method, MCSC \cite{Liu_2023_BMVC}, while being significantly simpler, and easier to adapt to use registration information.
For inference, we make predictions by averaging the logits from the two models.

\subsection{Learning from spatial registration}
\label{subsec:method-reg-overview}
% \pmh{say that although model remains 2d, we now consider each slice as situated in 3D space of its containing volume}
% \pmh{introduce notation for registrations---we denote the registration transform that maps points in volume x to y by $T_{xy}$}
% \pmh{possibly introduce a notation for the full set of pairwise registrations}
We now describe how our CCT-R incorporates registration information into the learning framework described in Section~\ref{subsec:method-prelims}. In CCT-R, spatial correspondences from registration serve as additional supervision, since points mapped together by an accurate registration transform share the same anatomical label across volumes.

We assume pairwise 3D registration transforms, either affine or deformable, are available between all volumes in \(V\); these can be calculated using any standard off-the-shelf method.
Although the segmentation model remains 2D, operating on individual slices, each slice is now considered within the 3D space of its original volume.
We define the set of registration transforms as \(T = \{T_{ij}\}_{i=1, j=1}^N\), where \(T_{ij}\) maps points from volume \(v_i\) to volume \(v_j\), and \(N\) is the total number of volumes.

% By utilizing \(T\), our framework focuses on two key aspects. 
Our CCT-R uses $T$ in two ways.
First, we go beyond cross-teaching, introducing a new loss that uses registration to transfer labels from labeled to unlabeled data
% function that encourages the model to extract segmentation knowledge from transforms of labeled and unlabeled pairs 
(Sec.~\ref{subsec:method-reg-sup}). 
Furthermore, traditional supervised contrastive learning typically relies on predicted logits, which can introduce errors. Our CCT-R mitigates this by using \(T\) to identify anatomically corresponding features across volumes, providing a complementary set of positives (Sec.~\ref{subsec:method-reg-contrast}). 
% By combining logit-based and registration-based supervision, we reduce the confirmation bias that can occur with standard supervised contrastive learning. \pmh{repeats start 3.3}
% These two aspects correspond to our proposed novel components, RSL and REPS, which are detailed in the following two sections.

\subsection{Registration supervision loss }
\label{subsec:method-reg-sup}

We use spatial transforms obtained by registration as an additional source of pseudo-labels to supervise the two models. Specifically, by transforming a point from an unlabeled volume to the corresponding point in a labeled volume, we can assume that these two points correspond to the same anatomical location. Thus, the label from the labeled volume can be used as supervision for the unlabeled slice. 
This provides much more accurate pseudo-labels early in training, and also helps to reduce the confirmation bias that can arise from cross-teaching.

Formally, we define a new loss $\mathcal{L}_{rs}$, that encourages each pixel to match the label of its corresponding location in the paired labeled volume:
\begin{align}
 \mathcal{L}_{rs}  &= - \frac1{M} \sum_{i=1} ^M \left(\mathcal{L}_{dice}( p^u_{i}, \,r^u_i) + \mathcal{L}_{ce}(p^u_i,\, r^u_i)\right),\label{l_rs} % \\
% r^u_i &=  y^l_i(T_{qj}(P_q))
\end{align}
where $p^u_i$ is the class probability map of the $i$\textsuperscript{th} unlabeled image $x^u_i$, and $r^u_i$ is a new registered label found by registration. $\mathcal{L}_{rs}$ is then added to the overall loss function (Eq.~\ref{total_loss}).

% \pmh{rewrite, as it's very hard to understand currently: say something like we assume that xui belongs to the jth unlabeled volume; we'll use the qth labeled volume to define the labels gui; use Tqj to map points of the slice from volume q to j; then retrieve the labels from that volume to create gui}

Assuming that the slice $x^u_i$ belongs to the unlabeled volume \( v^u_j \), we define the registered label $r^u_i$ by mapping the ground truth $y^l_i$ from the labeled volume \( v^l_q \):
\begin{equation}
r^u_i = T_{qj}(y^l_i),
\end{equation}
where $T_{qj}$ is the transform from \( v^u_q \) to \( v^l_j\). This transform aligns the label $y^l_i$ with the corresponding coordinates in the slice $x^u_i$, resulting in the $r^u_i$.
This greatly improves the model's learning performance (see Sec.~\ref{ablation:pseudo label}), especially in cases with minimal supervision (\eg only one labeled volume). 

\paragraph{Best registration selection strategy.}
In practice, registrations are often imperfect, particularly for complex anatomical regions such as the abdomen.
Moreover, the loss described in Eq.~\ref{l_rs} does not require every image to be paired with all others.
We therefore design a strategy to choose which registered pairs should be used.
Importantly, this strategy cannot rely on ground-truth labels, due to our semi-supervised setting.
%
% First the registration provides a new source of supervision, which significantly aids in learning from unlabeled data, particularly by improving the accuracy of pseudo-labels in the early stages when labeled data is sparse.
% To evaluate and select the best-matching pairs for generating pseudo-labels, we employ the Bidirectional Consistency Filtering strategy.
% \pmh{if we name the full set of registrations above, then here say we're finding a subset of them}\textcolor{red}{I wonder if it is necessary to define a subset to say that we select?}
% To determine which registrations to use, 
Specifically,
we measure the cycle-consistency of the transforms from $T$ (Sec.~\ref{subsec:method-reg-overview}) between two volumes, say \( v^u_j \) and \( v^l_q \).
We apply the forward transform \( T_{jq} \) (j-to-q) and the reverse transform \( T_{qj} \) (q-to-j) on volume \( v^u_j \):
\begin{equation}
\tilde{v}^u_j = T_{qj}(T_{jq}(v^u_j)).
\end{equation}
Ideally, \(\tilde{v}^u_j\) should be equal to the original volume \( v^u_j \), meaning the composition of forward and reverse transformations approximates the identity function.
We calculate the global similarity between $v^u_j$ and $\tilde{v}^u_j$ using both mutual information (MI) \cite{mattes2003pet} and root mean square error (RMSE), and use these to derive a composite score
\begin{equation}
S = w_\text{rmse} \cdot \text{RMSE} + w_\text{mi} \cdot \text{MI},
\end{equation}
where $w_\text{rmse}$ and $w_\text{mi}$ weight the importance of RMSE and MI, respectively. We then select the $v^l_q$ that minimizes this composite score to generate the best additional pseudo-label $r^u_i$ for the unlabeled slice $x^u_i$ in $v^u_j$.

\subsection{Registration-enhanced positive sampling}
\label{subsec:method-reg-contrast}

% We next show how to use registration to improve the supervised contrastive learning loss in Eq.~\ref{out}.
% For positive samples $a_p$, Eq.~\ref{out} uses pixels in the minibatch that have the same class as the anchor.
% We now augment the positive by including pixels from anatomically corresponding locations in multiple volumes as determined by registration.
% False or low-confidence pseudo-labels can compromise the accurate identification of organ boundaries \cite{wang2023space}.
% To maximize the benefits of contrastive learning, it is essential that feature sampling is as diverse, accurate, and representative as possible.
% Registration introduces an additional set of positive samples that do not rely on the current pseudo-labels and are distributed diversely across all volumes in the dataset, complementing those based on labels.
We next show how to use registration to improve the supervised contrastive learning loss in Eq.~\ref{out}.
\figureautorefname{}~\ref{fig:contrastive_learning} shows the shortcomings of standard positive sampling in comparison to our novel approach integrating registration.
Positives $a_p$ derived from (pseudo-)labels are sampled from any location within the same organ or class as shown in Eq.~\ref{out}.
%(subject to noise of pseudo-labels).
In contrast, registration-based positives correspond to the exact same anatomical location within the organ, albeit in different volumes or patients. Any noise in registration-based positives stems from registration inaccuracies and is independent of pseudo-label errors.
Therefore, we augment the set of positive samples by incorporating registration-based examples. This approach reduces the confirmation bias that can arise when learning only from pseudo-labels.
Assume the xyz coordinate of anchor $a_i$ in an image from volume $v_q$ is denoted by $p$. We use a registration transform to get the corresponding positive coordinates $p_j$ in $v_j$: 
%\pmh{do you look at (i.e. does j index over) all other volumes, or all labelled, or some subset? doesn't need to describe memory bank yet, but currently it's too vague}
\begin{equation}
p_j = T_{qj}(p),
\end{equation}
% = \mathbf{A}(p - c) + t + c. $\mathbf{A}$ is the transformation matrix, $\mathbf{c}$ is the center point, and $\mathbf{t}$ is the translation vector
where \( j \in \{1, 2, \dots, N\} \) and \( j \neq q \), i.e.~we consider all other training volumes in \(V\). Given the $p_j$, we extract the positive feature $a_{pj}^r$ from the corresponding feature maps.

Since our minibatch comprise 2D slices rather than full 3D volumes, there is only a small probability that the feature map containing a given registered point $p_j$ will in fact be available in the current minibatch.
We therefore build a memory bank $B$ to serve as a source of feature maps, which provides more diverse registered positive samples across different 3D volumes.
% $B$ is designed to store feature maps of 2D slices and updated for each mini-batch in a first-in, first-out (FIFO) order, while preserving a fixed size $K$.
% \pmh{this discussion of memory bank seems a bit short; is there any other details that are useful? in particular does the memory bank store entire slices? is there any strategy to decide which slices are kept / not?}
% \pmh{also maybe mention that the features are therefore slightly `out-of-date', c.f. BYOL}
The memory bank \( B \) stores feature maps of 2D slices.
For every slice in each mini-batch, new feature maps are added to \( B \).
If a slice is not yet in $B$, it is added; otherwise, the existing slice is updated with the new features.
Once \( B \) reaches its maximum capacity \( K \), the oldest slices are removed in a first-in, first-out (FIFO) order.
%Although the feature maps in \( B \) may be slightly ``out-of-date'' due to being accumulated across different mini-batches, 
This provides the model with a more diverse set of features from various 3D volumes. %, enhancing the training process.

The positive features $a_{pj}^r$ are averaged over the available $j$ indices that exist in the memory bank:
% Then the average is given by:
%
\begin{align}
a_{p}^r &= \frac{1}{\lvert J \rvert} \sum\limits_{j \in J} a_{pj},
\end{align}
where \( J \) represents the set of volume indices for which the feature point exists in the memory bank. Note that \( J \) is a subset of the total volume indices \( \{1, 2, \dots, N\} \).

Finally, we combine with the pseudo-label-supervised positive key $a_p^{l}$ from Eq.~\ref{eq:supd-positives} to give a single combined positive key $a_p$ for $a_i$:
\begin{equation}
a_p =  w_1 a_p^{l} + w_2 a_p^{r}.
\end{equation}
We use these positives in the contrastive loss Eq.~\ref{out}, but otherwise keep it unchanged.
% In this way, our DSSC module not only increases the diversity of positive samples but also lowers the risks of ignoring object boundaries as discussed in \cite{wang2023space}.  

\section{Experiments}

\paragraph{Datasets.}
We evaluate CCT-R using two challenging benchmark datasets.
\textbf{ACDC \cite{bernard2018deep}} comprises of 200 short-axis cardiac MR volumes from 100 cases, with segmentation masks provided for the left ventricle (LV), myocardium (Myo), and right ventricle (RV). 
% Following the data split and labeled case selection described in , 
We allocate 70 cases (1930 slices) for training, 10 for validation, and 20 for testing as in \cite{Luo2022}, and match their choice of labeled cases.
\textbf{Synapse \cite{landman2015miccai}} consists of abdominal CT volumes from 30 cases, with eight labeled organs: aorta, gallbladder, spleen, left kidney, right kidney, liver, pancreas, and stomach. As in \cite{Chen2021Transunet}, we use 18 cases (2212 slices) for training and 12 for testing.
We precomputed a composite pairwise registration (affine for ACDC and affine + B-spline deformable transformation for Synapse) for all training data using ITK \cite{lowekamp2013design, mccormick2014itk}.

\paragraph{Metrics.}
For quantitative evaluation, we use two widely-recognized metrics for 2D segmentation: Dice coefficient (DSC) and 95$\%$ Hausdorff Distance (HD).

\paragraph{Baselines.} 
We first compare with a registration baseline that is not learning-based---we use the transforms to propagate labels from the labeled training cases to the test images, similar to \cite{benson2014correction, fischl2002whole, lorenzo2002atlas}, selecting labeled cases with our BRS.
We also compare a joint registration and segmentation model, DeepAtlas \cite{xu2019deepatlas}; this learns registration from scratch simultaneously with segmentation. To stay consistent with our CCT-R, we reimplemented it using a 2D U-Net segmentation model.
We evaluate several recent S4 methods with the U-Net \cite{ronneberger2015u} backbone: Mean Teacher (MT) \cite{tarvainen2017mean}, Deep Co-Training (DCT) \cite{qiao2018deep}, Uncertainty Aware Mean Teacher (UAMT) \cite{yu2019uncertainty}, Interpolation Consistency Training (ICT) \cite{verma2022interpolation},  Cross Consistency Training (CCT) \cite{ouali2020semi}, Cross Pseudo Supervision (CPS) \cite{Chen2021a}, and Cross Teaching Supervision (CTS) \cite{Luo2022}, which like CCT-R uses Swin-UNet \cite{Cao2021} (Transformer) and U-Net backbones. 
In addition, we include the SOTA S4 method with contrastive learning, MCSC \cite{Liu_2023_BMVC}.
As a reference we also train the U-Net backbone from the S4 methods on only the labeled subset of cases (LS) without additional tricks.
We also include fully-supervised methods---the same U-Net trained under full supervision (FS), and the SOTA fully-supervised methods BATFormer \cite{Lin2023} (on ACDC) and nnFormer \cite{zhou2021nnformer} (on Synapse).
We retrain all baseline models using their recommended hyperparameters, and report the results from \cite{Luo2022} or our replication, whichever is better.
Furthermore, the results of all baselines are given in the appendix.
% % the best of our 
\paragraph{Implementation details.}
For all methods we use random cropping, random flipping and rotations to augment. All methods were trained until convergence, or up to 40,000 iterations. 
We precomputed a composite pairwise registration (affine for ACDC and affine + B-spline deformable transformation for Synapse) for all training data using ITK \cite{lowekamp2013design, mccormick2014itk}. 
We used the AdamW optimizer with a weight decay of $5\times10^{-4}$.  The learning rate followed a polynomial schedule, starting at $5\times10^{-4}$ for the U-Net and $1\times10^{-4}$ for the Swin-Unet. Our training batches consisted of 8 images for ACDC and 24 images for Synapse, evenly split between labeled and unlabeled. In the contrastive learning section, each ($H_*$) was composed of two linear layers, outputting 256 and 128 channels, respectively.
In Eq.~6, $w_{cps}$ is defined by a Gaussian warm-up function \cite{Luo2022}: $w_{cps}(i) = 0.1\cdot \exp\left(-5(1-i/t_\mathrm{total})^2\right)$, where $i$ is the index of the current training iteration and $t_\mathrm{total}$ is the total number of iterations, while $w_{cl}$ is set to a constant value of $10^{-3}$.
In Eq.~4, temperature $\tau = 0.1$. In REPS module, the bank size $K = (M+K)/5$.  
We implemented our method in PyTorch. All experiments were run on one RTX 3090 GPU.

\subsection{Comparison with Existing Methods} 

\begin{figure*}[t]
\begin{center}%
\includegraphics[width=0.49\linewidth]{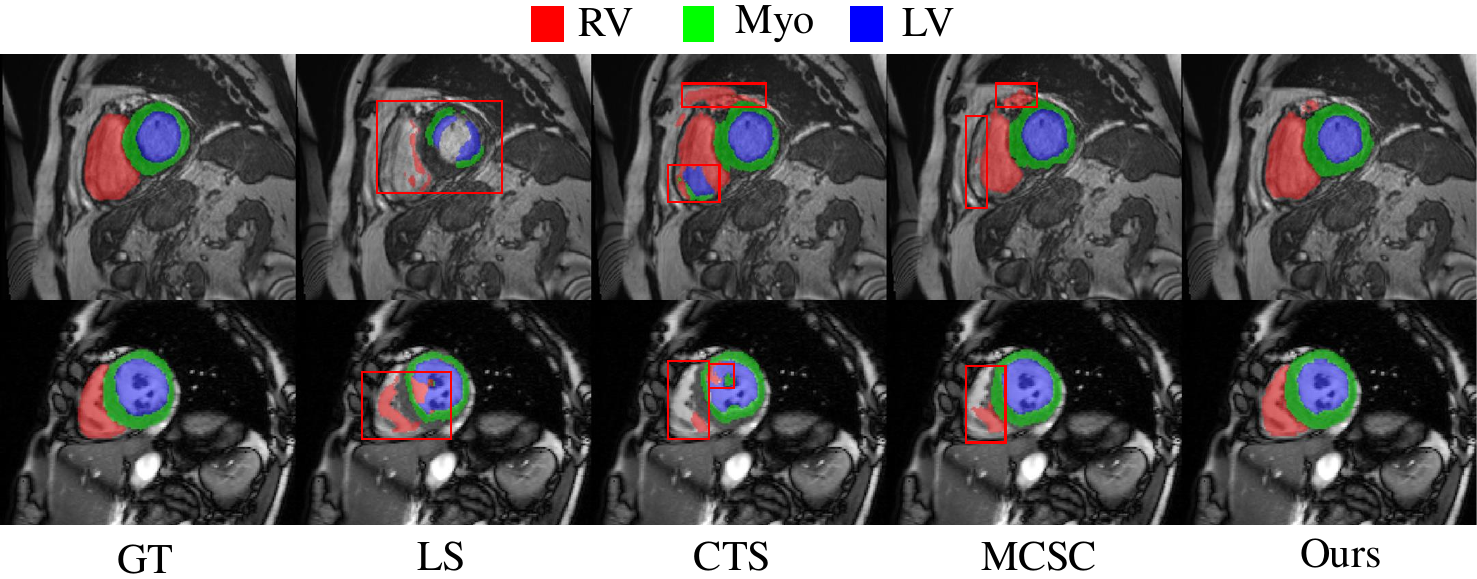}
\includegraphics[width=0.49\linewidth]{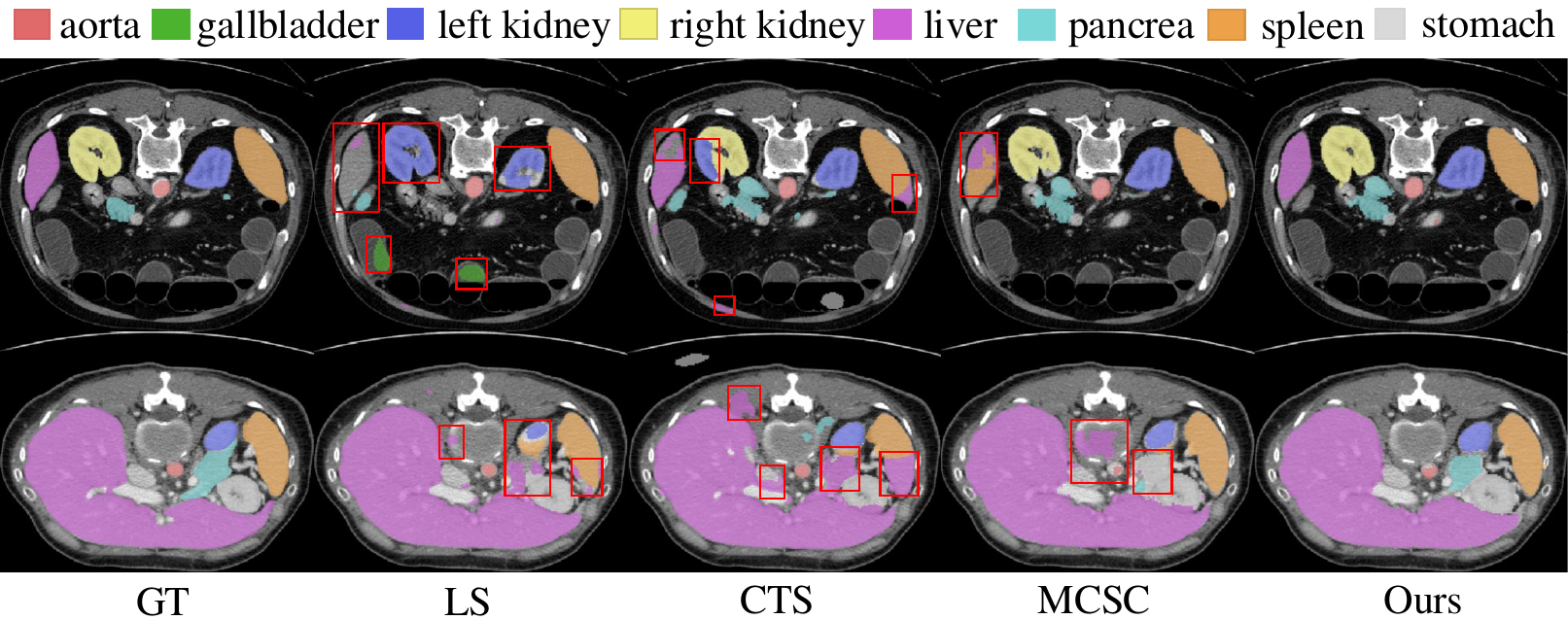}%
    \end{center}
    \vspace{-16pt}
\caption{Qualitative results from our CCT-R and baselines. \textbf{Left:} ACDC, trained on 3 labeled cases; \textbf{right:} Synapse, 2 labeled cases}
\label{fig:acdc-synapse}
% \vspace{-5pt}
\end{figure*}

\begin{table}[t]
\caption{Segmentation results on ACDC for our method and baselines, according to DSC ($\%$) and HD (mm).}
\vspace{-6pt}
  \centering
  \renewcommand*{\arraystretch}{1}
  \setlength{\tabcolsep}{3pt}
  %\begin{minipage}{1\linewidth}
    \centering

    \resizebox{\linewidth}{!}{
    \begin{tabular}{@{}llcccccccc@{}}
    \toprule
    \multirow{2}{*}{\textbf{Labeled \hspace{-4pt}}} & \multirow{2}{*}{\textbf{Methods}} & \multicolumn{2}{c}{\textbf{Mean}}  & \multicolumn{2}{c}{\textbf{Myo }}  & \multicolumn{2}{c}{\textbf{LV }} & \multicolumn{2}{c}{\textbf{RV }}
    \\\cmidrule(lr){3-4}\cmidrule(lr){5-6}\cmidrule(lr){7-8}\cmidrule(l){9-10}
    & & DSC$\uparrow$& HD$\downarrow$ & DSC$\uparrow$ & HD$\downarrow$ & DSC$\uparrow$ & HD$\downarrow$  & DSC$\uparrow$ & HD$\downarrow$ \\
    \midrule
   \multirow{2}{*}{ 70 (100$\%$)}
    & UNet-FS & 91.7  & 4.0  & 89.0  & 5.0 & 94.6  & 5.9  & 91.4 & 1.2  \\
    & BATFormer \cite{Lin2023} & 92.8  & 8.0    &90.26  & 6.8 & 96.3 & 5.9&  91.97  & 11.3  \\
     \midrule
     \multirow{13}{*}{ 7 (10$\%$)}&Reg. only (Aff) & 30.7 & 16.4 & 19.7 & 13.9 & 42.0 & 14.4 & 30.5 & 20.8 \\

    \cmidrule(r){2-10} 
    & DeepAtlas \cite{xu2019deepatlas}&  79.4 & 8.0  & 79.0 & 11.7  & 81.9 &\underline{ 3.2  }& 77.3 &  9.0   \\
    \cmidrule(r){2-10} 
     & UNet-LS  & 75.9   & 10.8 & 78.2  &8.6 & 85.5  &13.0& 63.9 &10.7  \\
      
     &MT \cite{tarvainen2017mean} & 80.9 &11.5 &79.1 &7.7&	86.1&13.4&77.6&13.3\\
    &DCT \cite{qiao2018deep} & 80.4 & 13.8  & 79.3& 10.7& 87.0 & 15.5 & 75.0 & 15.3\\
    &UAMT  \cite{yu2019uncertainty} & 81.1  & 11.2 &  80.1& 13.7& 87.1 & 18.1 & 77.6   & 14.7\\
    &ICT \cite{verma2022interpolation} & 82.4 & 7.2 & 81.5  & 7.8  & 87.6 & 10.6 & 78.2 & 3.2\\
    &CCT  \cite{ouali2020semi} & 84.0 & 6.6 & 82.3  & 5.4 & 88.6& 
9.4 & 81.0 & 5.1 \\
    &CPS \cite{Chen2021a} & 85.0& 6.6& 82.9   &6.6 & 88.0 &10.8 & 84.2 & 2.3\\
    &CTS  \cite{Luo2022} & 86.4 & 8.6 & 84.4 & 6.9  & 90.1 & 11.2 & 84.8 & 7.8\\
     & MCSC  \cite{Liu_2023_BMVC}  & \underline{89.4} &  \underline{2.3} &  {\bf87.6} & {\bf1.1}   & {\bf93.6} &  3.5 &  \underline{87.1} &  \underline{2.1}\\
    & Ours (Affine) & \bf{90.3} & \bf{1.6} & \underline{87.4 }& \underline{1.4} & \underline{92.7} & \bf{2.2} &\bf{90.9} & \bf{1.3} \\
     \midrule
     \multirow{13}{*}{ 3 (5$\%$)}&Reg. only (Aff) & 32.0 & 17.8 & 18.0 & 15.7 & 43.9 & 16.0 & 34.0 & 21.7 \\
    \cmidrule(r){2-10} 
     & DeepAtlas \cite{xu2019deepatlas}&  59.0 & 8.6 & 62.8& \underline{5.4} & 67.8 & \underline{7.7} & 46.4 & 12.6 \\
      \cmidrule(r){2-10} 
     & UNet-LS  & 51.2   & 31.2 & 54.8  & 24.4 & 61.8  &24.3& 37.0 & 44.4  \\
     &MT \cite{tarvainen2017mean}  & 56.6 &34.5& 58.6 &23.1&	70.9&26.3 &40.3&53.9\\
    &DCT \cite{qiao2018deep}  & 58.2 & 26.4 & 61.7 & 20.3 & 71.7& 27.3 & 41.3 & 31.7\\
    &UAMT  \cite{yu2019uncertainty}  & 61.0  & 25.8 & 61.5   & 19.3&  70.7& 22.6& 50.8 & 35.4\\
    &ICT \cite{verma2022interpolation} & 58.1 & 22.8 & 62.0  & 20.4  & 67.3 & 24.1 & 44.8 & 23.8\\
    &CCT  \cite{ouali2020semi}  & 58.6 & 27.9 & 64.7  & 22.4& 70.4& 27.1 & 40.8 &34.2 \\
    &CPS \cite{Chen2021a}   & 60.3& 25.5& 65.2   &18.3 & 	72.0 &22.2 & 43.8 & 35.8 \\
    &CTS \cite{Luo2022}   & 65.6 & 16.2 & 62.8 &11.5  & 76.3 & 15.7 & 57.7 & 21.4\\
    & MCSC  \cite{Liu_2023_BMVC} & \underline{73.6 } & \underline{10.5}  & \underline{70.0} & 8.8 & \underline{79.2}& 14.9& \underline{71.7} & \underline{7.8}\\
    & Ours (Affine) & \bf{85.7 }& \bf{2.0 }& \bf{83.8} & \bf{1.4 }&\bf{ 89.9 }&\bf{ 2.4 }& \bf{83.5 }& \bf{2.1} \\

    \midrule
    \multirow{7}{*}{ 1 (1.4\%) } &Reg. only (Aff) & 23.4 & 19.7 & 13.6 & 18.7 & 31.6 & 19.0 & 25.1 & 21.4 \\ 
    \cmidrule(r){2-10} 
     & DeepAtlas \cite{xu2019deepatlas}& 40.4 & 18.5 & 42.2 & 11.7 & 34.7 & 29.2 & 44.4 & \underline{14.6}    \\
      \cmidrule(r){2-10} 
    & UNet-LS  &  26.4   & 60.1 & 26.3  & 51.2 & 28.3  & 52.0& 24.6 & 77.0  \\

    &CTS  \cite{Luo2022}  & 46.8 &  36.3  & 55.1 & \underline{5.5}&	64.8 & {\bf4.1} &20.5 & 99.4\\
     &MCSC  \cite{Liu_2023_BMVC} & \underline{58.6 } &\underline{ 31.2}& \underline{64.2} &13.3&	\underline{78.1} &  12.2 & \underline{33.5} & 68.1\\
       & Ours (Affine) &\bf{ 80.4 }&\bf{ 3.5} & \bf{78.3 }& \bf{3.2} & \bf{83.6} &  \underline{4.3} & \bf{79.3} &\bf{ 2.9} \\
    \bottomrule 
    \multicolumn{10}{r}{\scriptsize \textbf{Best} is bold, \underline{Second Best} is underlined. \hfill}
    \end{tabular}
    }
    \label{tab:acdc}
\end{table}

\paragraph{ACDC.}
Table \ref{tab:acdc} presents quantitative results from our CCT-R and baselines, under three different levels of supervision (7, 3, and 1 labeled cases). When trained on 7 labeled cases (10$\%$),  significantly outperforms the baseline CTS, with more than a 4$\%$ improvement in DSC and a reduction of 7~mm in HD. 
% Specifically,  achieves a 1.6mm HD, outperforming CTS's 8.6mm, and a DSC of 90.3$\%$, compared to CTS's 86.4$\%$.
%It's effectiveness becomes even more pronounced when working with even less labeled data. 
With just 5$\%$ of labeled data (3 cases), our CCT-R surpasses CTS and SOTA MCSC by an impressive margin of 20$\%$ and 12$\%$ in DSC and reduction of 14~mm and 8.5~mm in HD, respectively.
%Even as the amount of labeled data decreases from 10$\%$ to 5$\%$,  continues to surpass CTS by a wide margin, with an approximately 18$\%$ improvement in DSC and a substantial reduction in HD for the right ventricle (RV)—a minority and challenging class. 
%  achieves a DSC of 85.7$\%$ and HD of 2.0mm, compared to SOTA MCSC's 73.6$\%$ DSC and 10.5mm HD under the same conditions.
When the supervision is reduced to one labeled case, our approach outperforms the SOTA by an even larger margin (DSC of 80.4 vs. 58.6 for MCSC), highlighting its robustness in scenarios with extremely limited labeled data. 
%In addition, the joint registration and segmentation method,
DeepAtlas, a joint registration and segmentation method, underperforms. This may be due to its lack of advanced S4 techniques, and its online learning of registration, which means registrations are inaccurate early in training and provide poor guidance for segmentation.
Qualitative results in 
\figureautorefname{}~\ref{fig:acdc-synapse} (left) further illustrate the superiority of CCT-R, showing more accurate segmentation with fewer under-segmented regions for the RV (bottom) and fewer false positives (top) compared to CTS.
%Overall,  shows a significant improvement in performance over SOTA methods, especially in scenarios with limited labeled data, consistently achieving better accuracy and robustness across different levels of supervision.
% In the supplementary (Sec.~S3) we also show that CCT-R outperforms CTS combined with other contrastive losses.

\begin{table}[t]
	\caption{Segmentation results on Synapse for ours method and baselines, according to DSC ($\%$) and HD (mm).}
 \vspace{-6pt}
	% The performance is reported by DSC (\%) and HD (\%), as well as the DSC value of each types of organs.
	\renewcommand*{\arraystretch}{1}
	\setlength{\tabcolsep}{2pt}
	\centering
	\resizebox{\linewidth}{!}{
		\begin{tabular}{@{}c c c c c c c c c c c c@{}}
			\toprule
			\textbf{Labeled} & \textbf{Methods} & DSC$\uparrow$ & HD$\downarrow$ & Aorta & Gallb & Kid$\_$L & Kid$\_$R & Liver & Pancr & Spleen & Stom \\
			\midrule
			\multirow{2}{*}{18(100$\%$)} &  UNet-FS  & 75.6 & 42.3 & 88.8 & 56.1 & 78.9 & 72.6 & 91.9 & 55.8 & 85.8 & 74.7 \\
			&  nnFormer & 86.6 & 10.6 & 92.0 & 70.2 & 86.6 & 86.3 & 96.8 & 83.4 & 90.5 & 86.8 \\
			\midrule
			\multirow{12}{*}{4(20$\%$)} & Reg. only (Affine) & 27.0 & 39.6 & 16.0 & 7.5 & 36.4 & 33.0 & 56.8 & 13.1 & 28.5 & 25.1  \\
                & Reg. only (Aff+Def) &   32.5  & 36.5  & 29.7  & 4.8  & 36.5 & 29.4  & 65.5 & 14.2  & 48.0  & 31.7   \\
                \cmidrule(r){2-12}
                  & DeepAtlas \cite{xu2019deepatlas}& 56.1&85.3&69.2&\underline{43.3}&50.8&55.2&88.8&30.5&62.7&48.0 \\
                  \cmidrule(r){2-12}

			& UNet-LS & 47.2 & 122.3 & 67.6 & 29.7 & 47.2 & 50.7 & 79.1 & 25.2 & 56.8 & 21.5 \\
			& UAMT \cite{yu2019uncertainty}  & 51.9 & 69.3 & 75.3 & 33.4 & 55.3 & 40.8 & 82.6 & 27.5 & 55.9 & 44.7 \\
			% & ICT \cite{verma2022interpolation} & 57.5 & 79.3 & 74.2 & 36.6 & 58.3 & 51.7 & 86.7 & 34.7 & 66.2 & 51.6 \\
			% & CCT \cite{ouali2020semi} & 51.4 & 102.9 & 71.8 & 31.2 & 52.0 & 50.1 & 83.0 & 32.5 & 65.5 & 25.2 \\
			& CPS \cite{Chen2021a} & 57.9 & 62.6 & 75.6 & 41.4 & 60.1 & 53.0 & 88.2 & 26.2 & 69.6 & 48.9 \\
			% & CTS(CNN) & 62.8 & 91.2 & 79.1 & 30.3 & 69.2 & 66.3 & 84.8 & 42.4 & 76.5 & 53.5 \\
			&CTS\cite{Luo2022} & 64.0 &  56.4 & \underline{79.9} & 38.9 & 66.3 & 63.5 & 86.1 & 41.9 & 75.3 &60.4 \\
                & MCSC \cite{Liu_2023_BMVC}& 68.5 & 24.8 & 76.3 & \bf{44.4} & \bf{73.4} &  \bf{72.3} & 91.8 &46.9 & 79.9 & 62.9 \\
                & Ours (Affine)&\underline{70.0}& \underline{23.2}& 79.8 &34.5& 71.0 & \underline{70.7} &\underline{92.8} & \bf{49.6} &\underline{87.4} & \bf{74.4} \\
		    & Ours (Affine+Deform)  & \bf{71.4} & \bf{21.1} & \bf{80.4} & 42.3 & \underline{73.0} & 70.0 & \bf{93.7} & \underline{49.4} & \bf{87.9} & \underline{74.2} \\
			\midrule
			\multirow{12}{*}{2(10$\%$)} & Reg. only (Affine) &25.4&36.8&17.5 & 3.5 & 32.7 & 27.5 &53.4 & 12.6 & 33.4 &22.5  \\
                & Reg. only (Aff+Def) &  29.1 & 44.0 & 27.2 & 11.3 & 28.6 & 26.5 & 66.4 & 12.7 & 29.7 & 30.3    \\
                \cmidrule(r){2-12}
                & DeepAtlas \cite{xu2019deepatlas}& 44.0& 67.1& 68.0  & 24.9 & 37.9  & 46.0 & 82.7  & 18.4 & 44.2  &30.6  \\
                  \cmidrule(r){2-12}
			& UNet-LS & 45.2 & 55.6 & 66.4 & 27.2 & 46.0 & 48.0 & 82.6 & 18.2 & 39.9 & 33.4 \\
			& UAMT \cite{yu2019uncertainty} & 49.5 & 62.6 & 71.3 & 21.1 & 62.6 & 51.4 & 79.3 & 22.8 & 58.2 & 29.0 \\
			% & ICT \cite{verma2022interpolation} & 49.0 & 59.9 & 68.9 & 19.9 & 52.5 & 52.2 & 83.7 & 25.4 & 53.2 & 36.0 \\
			% & CCT \cite{ouali2020semi} & 46.9 & 58.2 & 66.0 & 26.6 & 53.4 & 41.0 & 82.9 & 21.2 & 48.7 & 35.6 \\
			& CPS \cite{Chen2021a} & 48.8 & 65.6 & 70.9 & 21.3 & 58.0 & 45.1 & 80.7 & 23.5 & 58.0 & 32.7 \\
			% & CTS(CNN) & 51.3 & 64.4 & 75.3 & 17.3 & 58.8 & 65.4 & 78.8 & 20.8 & 71.4 & 22.5 \\
			& CTS \cite{Luo2022}  & 55.2 & 45.4 & 71.5 & 25.6 & 62.6 & 67.5 & 78.2 & 26.3 & 75.9 & 34.3 \\
                & MCSC \cite{Liu_2023_BMVC}&  61.1 &  32.6 & 73.9 & 26.4 &  69.9 & 72.7 & 90.0 & 33.2 & 79.4 & 43.0  \\
                & Ours (Affine)& \underline{65.1}	&\underline{22.5}&\underline{75.7}&	\underline{28.4}	&\underline{74.5}&\bf{75.0}&\underline{91.8}&\underline{38.0}&\bf{82.3}&\underline{55.1} \\
			& Ours (Affine+Deform)  & \bf{66.5} & \bf{19.7} & \bf{77.6} & \bf{34.4} & \bf{75.1} & \underline{74.2} & \bf{92.6} & \bf{39.5} & \underline{82.1} & \bf{56.1} \\
			\midrule
			\multirow{12}{*}{1(5$\%$)} & Reg. only (Affine) & 26.4  & 45.0  & 16.3  & 6.6  & 35.8 & 32.8  & 53.5 & 14.4  & 28.7  & 22.7  \\
                & Reg. only (Aff+Def) &  27.4  & 52.2  & 26.4  & 11.3  & 30.5 & 27.1  & 61.6 & 12.8  & 26.3  & 23.6   \\
                \cmidrule(r){2-12}
            & DeepAtlas \cite{xu2019deepatlas}&  16.1 & 72.3 & 18.4 & \underline{14.9}  &1.2  & 10.1  & 57.1 & 0.6 &14.4 &12.2    \\
                  \cmidrule(r){2-12}
			& UNet-LS & 13.7 & 116.5 & 11.6 & \bf{17.8} & 0.8 & 1.8 & 56.9 & 0.1 & 8.7 & 11.6 \\
			& UAMT\cite{yu2019uncertainty}  & 10.7 & 90.2 & 8.0 & 9.3 & 0.3 & 8.1 & 31.7 & 1.1 & 13.1 & 14.3 \\
			% & ICT\cite{verma2022interpolation} & 15.9 & 82.3 & 13.8 & 11.9 & 0.3 & 2.7 & 70.5 & 0.8 & 16.4 & 10.6 \\
			% & CCT\cite{ouali2020semi} & 11.7 & 107.5 & 10.0 & 13.0 & 0.1 & 1.9 & 47.5 & 3.7 & 8.0 & 9.3 \\
			& CPS \cite{Chen2021a} & 15.0 & 123.5 & 19.6 & 9.6 & 5.6 & 6.9 & 59.4 & 2.3 & 9.4 & 7.2 \\
			% & CTS(CNN) & 25.1 & 95.4 & 47.8 & 0 & 3.1 & 3.1 & 64.5 & 10.5 & 49.0 & 22.7 \\
			& CTS \cite{Luo2022}& 26.3 & 96.5 & 44.6 & 4.0 & 11.2 & 5.5 & 60.3 & 9.6 & 54.1 & 21.2 \\
                & MCSC \cite{Liu_2023_BMVC}&34.0 & 53.8 & 50.9 & 13.0 & 17.6& 54.6 & 64.3 & 5.5 & 43.1 & 23.5 \\ 
			& Ours (Affine)  & \underline{43.4} & \underline{40.8} & \underline{62.5} & 13.3 & \underline{17.9} & \bf{71.0} & \bf{77.0} & \bf{11.4} & \underline{65.4} & \bf{28.7} \\
         & Ours (Affine+Deform)& \bf{47.6}	&\bf{38.4}&\bf{65.5}&9.3&\bf{50.6}&\underline{70.2}&\underline{72.7}&	\underline{11.1}&\bf{73.9}	&\underline{27.8} \\
			\bottomrule
			\multicolumn{12}{r}{\scriptsize \textbf{Best} is bold, \underline{Second Best} is underlined. \hfill}
		\end{tabular}
	}
	\vspace{-5pt}
	\label{tab:synapse}
\end{table}

\paragraph{Synapse.}
We evaluate performance on the Synapse dataset using 4, 2, and 1 labeled cases. Although Synapse is more challenging than ACDC due to greater class imbalance and anatomical variability,
CCT-R demonstrates even larger improvements than on ACDC (Table~\ref{tab:synapse}). With 4 labeled cases, DSC increases from 64.0$\%$ to 71.4$\%$, outperforming CTS by 7.4$\%$ and MCSC by 2.9$\%$.
% Notably, with 2 labeled cases, CCT-R surpasses SOTA MCSC by a considerable margin (+5.4$\%$ and -12.9mm).
Even with just one labeled case, CCT-R still excels at segmenting challenging small organs like the aorta, kidney, and pancreas, where others struggle. It significantly outperforms MCSC, improving the mean DSC by 13.6$\%$ and reducing HD by 15.4~mm.
This robustness to extreme class imbalance and limited supervision emphasizes the value of registration information.
Furthermore, our approach is robust across varying registration qualities.
Even with simpler affine registrations, inaccurate for complex abdominal anatomy, it significantly improves segmentation (\textit{Ours (Affine)} rows) over not using registration, though results are better still with deformable transforms (\textit{Ours (Affine+Deform)}).
\figureautorefname~\ref{fig:acdc-synapse} (right) shows CCT-R accurately segments small structures like the gallbladder and pancreas, often missed or over-segmented by LS and CTS. Our approach also correctly identifies the spleen and distinguishes it from the liver, a common error in other methods. It also provides more precise segmentation of the liver and stomach, significantly outperforming MCSC. This figure shows the robustness in handling challenging, imbalanced datasets.

\paragraph{Segmentation via registration only.}
We also test whether simply propagating labels based on either affine or deformable registration achieves adequate segmentation performance (\textit{Reg.only} rows in Tables \ref{tab:acdc} \& \ref{tab:synapse}).
We see this performs substantially worse than the learning-based methods. 
%\pmh{quote a couple of numbers}
%When the setting is more than one labeled image, we apply our BRS strategy to select the best transform for generating label for each test image, ensuring a fair comparison.

%As shown in Table~\ref{tab:registration}, the registration approach on its own falls short of providing accurate segmentation and lags behind . 

\subsection{Benefit of Our Registration-Based Modules Applied on Different Baselines}
\label{subsec:diffbaseline}

Our main experiments build on CTS; however to show the wide applicability of our approach, we measure performance when it is integrated with alternative SSL baselines (Table~\ref{tab:benefit}).
We include UAMT \cite{yu2019uncertainty}, a classic teacher-student framework with two U-Nets, CPS~\cite{Chen2021a}, a student-student framework with two cross-teaching U-Nets, and CTS~\cite{Luo2022}, which improves CPS by replacing one of the U-Nets with Swin-UNet.
With each baseline, we measure the benefit of adding RSL only, and RSL in conjunction with contrastive learning and registration-based positive selection (\textit{SCL + REPS} row).
Our registration-derived modules boost all baselines. Enhanced UAMT approaches CTS performance, while improved CPS surpasses CTS by 4$\%$ on DSC. CTS with our modules remains the top performer.
%\textcolor{red}{fixed}
% \pmh{we should say sth specific to how well we assist different frameworks (e.g. UAMT with our bits becomes between than CPS etc; best baseline CTS is still best with ours; etc)}
% We attribute this large improvement to the fact that registration helps the S4 framework to extract richer semantic information from unlabeled data. Furthermore, registration-guided sampling prevents the model from over-reliance on pseudo on potentially erroneous pseudo-labels by integrating multiple sources of supervision.

\begin{table}[t]
\scriptsize
\centering
  \renewcommand*{\arraystretch}{0.75}
  \setlength{\tabcolsep}{1pt}
\caption{Benefit of our modules combined with different baselines, on Synapse with 10$\%$ labeled data.}
\vspace{-6pt}
\scalebox{1}{
\begin{tabular}{@{}lccccccc@{}}
    \toprule
     & \multicolumn{2}{c}{\textbf{UAMT}~\cite{yu2019uncertainty}} & \multicolumn{2}{c}{\textbf{CPS}~\cite{Chen2021a}} & \multicolumn{2}{c}{\textbf{CTS}~\cite{Luo2022}}
        \\\cmidrule(r){2-3}\cmidrule(lr){4-5}\cmidrule(lr){6-7}
     & DSC$\uparrow$ & HD$\downarrow$ & DSC$\uparrow$ & HD$\downarrow$ & DSC$\uparrow$ & HD$\downarrow$ \\
    \midrule
     Baselines & 49.5 & 62.6  &48.8 & 65.6 & 55.2 & 45.4  \\
    \midrule
     + RSL &52.3 & 60.3   & 57.3 & 42.4   &   65.4   &   28.5 \\ 
     + RSL + SCL + REPS &  54.6  & 55.6  &59.1 &37.5  &   66.5  &  19.7      \\ 
    \bottomrule
\end{tabular}
	}
\label{tab:benefit}
\end{table}

\subsection{Comparison with Alternative Supervised Contrastive Learning Losses}

In Table~\ref{tab:contrastive}, we compare our proposed approach with the state-of-the-art contrastive S4 method MCSC \cite{Liu_2023_BMVC}, and with incorporating other recent patch-level and slice-level contrastive learning techniques (GLCL~\cite{Hu2021} and ReCo \cite{liu2022bootstrapping}) into CTS.
% Given that GLCL and ReCo were initially developed for general contrastive learning rather than specifically for S4, we have reintegrated them into the CTS cross-teaching framework to provide an equitable comparison with our approach.
While all the contrastive losses improve on vanilla CTS, our CCT-R 
% demonstrates a superior ability to capture feature relationships across the entire dataset, resulting in 
achieves higher segmentation accuracy on nearly all datasets and labelling rates.

\begin{table}[t]
\centering
  \renewcommand*{\arraystretch}{1}
  \setlength{\tabcolsep}{1pt}
% \caption{Benefit of our method combined with different baselines, on Synapse with 20$\%$ labeled data.}
\caption{Comparisons with SoTA contrastive learning methods combined with CTS, on ACDC and Synapse.}
\scalebox{0.61}{
\begin{tabular}{@{}llccccccccc@{}}
    \toprule
      \multicolumn{2}{l}{\multirow{2}{*}{\textbf{Contrastive learning method}}} & \multicolumn{2}{c}{\textbf{ACDC~3 (5 $\%$)}} & \multicolumn{2}{c}{\textbf{/~1 (1.4 $\%$)}} & \multicolumn{2}{c}{\textbf{Synapse 4 (20 $\%$)}}& \multicolumn{2}{c}{\textbf{/~2 (10$\%$)}}
        \\\cmidrule(r){3-4}\cmidrule(r){5-6}\cmidrule(r){7-8}\cmidrule(r){9-10}
    & & DSC$\uparrow$ & HD$\downarrow$ & DSC$\uparrow$ & HD$\downarrow$ & DSC$\uparrow$ & HD$\downarrow$ & DSC$\uparrow$ & HD$\downarrow$\\
    
    \midrule
    \multirow{2}{*}{Patch-level}& GLCL \cite{Hu2021} (MICCAI'21) & 71.7  & 3.8 & 47.4 &  35.8   & 67.7   & 42.6  &  59.7  &34.6 \\ 
     &MCSC \cite{Liu_2023_BMVC} (BMVC'23) & 73.6  & 10.5 & 58.6 & 31.2     &    68.5  & 24.8 & 61.1   & 32.6 \\ 
     \midrule
    \multirow{2}{*}{Slice-level}&  ReCo \cite{liu2022bootstrapping} (ICLR'22) & 70.2  & 6.1 & 48.3 & 33.5      &  68.3   & 25.9 &   60.4 & 20.7\\ 
     & Ours  &  {\bf85.4}  & {\bf2.6}& {\bf80.0}&  {\bf4.2}&  {\bf71.4}  & {\bf21.1} & {\bf66.5} &   {\bf19.7} \\ 
     \midrule
    \multicolumn{2}{l}{None (Vanilla CTS~\cite{Luo2022})} & 65.6 & 16.2& 46.8 & 36.3  & 64.0 &56.4 & 57.2 & 45.7    \\
    \bottomrule
    \multicolumn{10}{r}{\scriptsize \textbf{Best} is bold. \hfill}
\end{tabular}

	}
\label{tab:contrastive}
\end{table}

\subsection{Ablation Studies and Analysis}  

\begin{table}[t]
\caption{Ablation study for the primary components of our CCT-R.
SCL: typical supervised local contrastive loss. RSL: registration supervision loss. BRS: best registration selection strategy for registered labels $r^u$. REPS: registration-enhanced positive sampling module (using positives from registration in SCL).} 
\vspace{-6pt}
\centering
\scalebox{0.7}
{
\begin{tabular}{@{}c c c c c c c c@{}}
				\toprule
				\multirow{2}{*}{SCL}&  \multirow{2}{*}{RSL }&  \multirow{2}{*}{BRS }&  \multirow{2}{*}{REPS }& \multicolumn{2}{c}{\textbf{~1 (5$\%$)}}   &\multicolumn{2}{c}{\textbf{~2 (10$\%$)}} 
         \\
         \cmidrule(lr){5-6}\cmidrule(lr){7-8}
				&  & & & DSC$\uparrow$ & HD$\downarrow$  & DSC$\uparrow$ & HD$\downarrow$ \\				
				\midrule
				 &  & & & 26.3 & 96.5&    55.2& 45.4 \\
			&\cmark&  &  & 29.0& 46.9  & 64.2&33.9\\
                &\cmark& \cmark &   &  -- & -- & 65.4&28.5 \\              
				\cmark & & &  &  27.5&59.8 &    63.1 & 29.1  \\
                \cmark& \cmark & \cmark & &   28.1 & 53.9 &  64.8 & 20.6\\	
    \cmark & & &  \cmark   & 31.4 & 55.2&  63.9 & 29.7   \\
    \cmark & \cmark& \cmark & \cmark  &  47.6 &38.4  & 66.5 &19.7  \\
\bottomrule
\end{tabular}}\\
\label{tab:ablation:components}
\end{table}

We conduct an ablation study on Synapse, measuring the importance of various aspects of our proposed CCT-R (Table~\ref{tab:ablation:components}). CTS, as our baseline, achieves Dice of 26.3$\%$ and 55.2$\%$ for one and two labeled cases respectively (top row).
% (CTS in Table~\ref{tab:synapse}), 
Our registration supervision loss (RSL) improves the baseline by +2.7$\%$ and 9.0$\%$. 
%which treats transforms as an additional source of pseudo-label to supervise the S4 framework, brings an improvement of +2.7$\%$ and 9.0$\%$ over the baseline. 
The best registration selection strategy (BRS), which is only applicable for two or more labeled cases, further boosts performance by an additional +1.2$\%$ in DSC and reduces HD by -5.4~mm. %Note that previous row chooses registration randomly. \textcolor{red}{fixed}
Adding a standard supervised local contrastive learning (SCL) improves the baseline by +1.2$\%$ and 7.9$\%$ respectively even without registration; also incorporating RSL gives further improvements of $0.6\%$ and $1.7\%$, indicating that contrastive learning and RSL are complementary strategies.
The registration-enhanced positive sampling (REPS), which mitigates bias towards single pseudo-label supervision in SCL, yields significant improvements: a +3.9$\%$ DSC and -4.6~mm HD for one labeled case and  +0.8$\%$ for two labeled cases versus just SCL.
Lastly, when combining all components, our full method achieves substantial Dice score improvement compared to the CTS baseline of 21.3$\%$ for 1 labeled case (from 26.3$\%$ to 47.6$\%$) and 11.3$\%$ for 2 labeled cases (from 55.2$\%$ to 66.5$\%$).
% These improvements can be attributed to the additional supervision provided by registration, which increases the accuracy of pseudo-labels on unlabeled data and offers more diverse and reliable dual-supervision in contrastive learning.

\begin{figure}[t]
\begin{center}
\includegraphics[width=\linewidth]{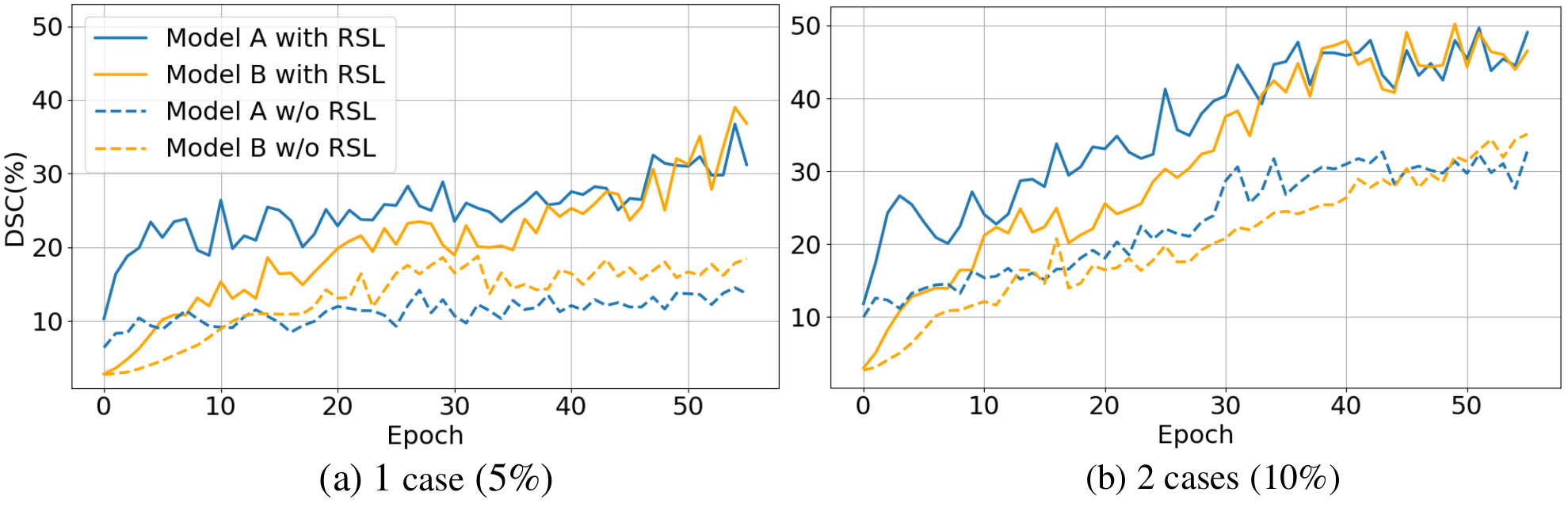}
    \end{center}
\vspace{-16pt}
\caption{DSC of pseudo-labels from two models on unlabeled data during the early training stages, for Synapse (a) 1 labeled case, and (b) 2 labeled cases.}
\label{fig:DSC_Pseudo_Labels}
% \vspace{-20 pt}
\end{figure}

\paragraph{Analysing the quality of pseudo-labels.} 
\label{ablation:pseudo label}
We measured the DSC of pseudo-labels predicted for unlabeled training data and used for cross-teaching, illustrating the noisiness of pseudo-labels and demonstrating how the proposed RSL mitigates this issue.
\figureautorefname{}~\ref{fig:DSC_Pseudo_Labels} shows that early in training, cross-teaching models without RSL (dashed lines) yield suboptimal results due to the insufficient training. This limitation persists even in later training stages, as the model struggles to generalize and often converges to local optima, especially in the 5$\%$ labeled setting. In contrast, the supervision provided by registrations, RSL, offers consistent and reliable guidance throughout the training process (solid lines), significantly mitigating these issues and enabling more effective learning from limited data.

\section{Conclusion}
We have introduced CCT-R, a registration-guided method for semi-supervised medical image segmentation.
This builds on cross-teaching methods, and improves segmentation via two novel modules: the Registration Supervision Loss and Registration-Enhanced Positive Sampling module. 
The RSL uses segmentation knowledge derived from transforms between labeled and unlabeled volume pairs, providing an additional source of supervision for the models.
With the REPS, supervised contrastive learning can sample anatomically-corresponding positives across volumes.
Without introducing extra training parameters, CCT-R achieves the new SOTA on popular S4 benchmarks. 

%%%%%%%%% REFERENCES

{\small
\bibliographystyle{ieee_fullname}
\bibliography{final}
}

\clearpage
\appendix
\section{Additional Results}
Here we show extended versions of Table~\ref{tab:acdc} and Table~\ref{tab:synapse} in the main paper as Table~\ref{tab:full_acdc} and Table \ref{tab:full_synapse}. In these extended tables, we provide additional comparisons by separately evaluating the performance of the two branches (CNN and Transformer) of our CCT-R (whereas in the main paper we use the mean of their logits); we also give results for all baselines under three different settings on both datasets.
 % summarizes ACDC segmentation results of our CCT-R and all baselines on 7, 3 and 1 labeled cases. 
 % shows the segmentation results of all methods on Synapse dataset under three settings (4, 2 and 1 labeled cases).
It can be seen that on the ACDC dataset, the performance of CCT-R's CNN and Transformer branches is quite similar. However, on the more challenging Synapse dataset, the Transformer outperforms the CNN, likely due to its superior ability to capture long-range dependencies, which allows it to better handle the relationships between large and small organs.

\begin{table*}
\caption{Segmentation results on ACDC for our method CCT-R and baselines, according to DSC($\%$) and HD(mm) for organs.}
  \centering
  \renewcommand*{\arraystretch}{1}
  \setlength{\tabcolsep}{3pt}
  % \begin{minipage}{1\linewidth}
    \centering
    {\small
    \begin{tabular}{@{}llcccccccc@{}}
    \toprule
    \multirow{2}{*}{\textbf{Labeled \hspace{-4pt}}} & \multirow{2}{*}{\textbf{Methods}} & \multicolumn{2}{c}{\textbf{Mean}}  & \multicolumn{2}{c}{\textbf{Myo }}  & \multicolumn{2}{c}{\textbf{LV }} & \multicolumn{2}{c}{\textbf{RV }}
    \\\cmidrule(lr){3-4}\cmidrule(lr){5-6}\cmidrule(lr){7-8}\cmidrule(l){9-10}
    & & DSC$\uparrow$& HD$\downarrow$ & DSC$\uparrow$ & HD$\downarrow$ & DSC$\uparrow$ & HD$\downarrow$  & DSC$\uparrow$ & HD$\downarrow$ \\
    \midrule
   \multirow{2}{*}{ 70 (100$\%$)}
    & UNet-FS & 91.7  & 4.0  & 89.0  & 5.0 & 94.6  & 5.9  & 91.4 & 1.2  \\
    & BATFormer \cite{Lin2023} & 92.8  & 8.0    &90.26  & 6.8 & 96.3 & 5.9&  91.97  & 11.3  \\
     \midrule
     \multirow{14}{*}{ 7 (10$\%$)}&Reg. only (Aff) & 30.7 & 16.4 & 19.7 & 13.9 & 42.0 & 14.4 & 30.5 & 20.8 \\

    \cmidrule(r){2-10} 
    & DeepAtlas \cite{xu2019deepatlas}&  79.4 & 8.0  & 79.0 & 11.7  & 81.9 & 3.2 & 77.3 &  9.0   \\
    \cmidrule(r){2-10} 
     & UNet-LS  & 75.9   & 10.8 & 78.2  &8.6 & 85.5  &13.0& 63.9 &10.7  \\
      
     &MT \cite{tarvainen2017mean} & 80.9 &11.5 &79.1 &7.7&	86.1&13.4&77.6&13.3\\
    &DCT \cite{qiao2018deep} & 80.4 & 13.8  & 79.3& 10.7& 87.0 & 15.5 & 75.0 & 15.3\\
    &UAMT  \cite{yu2019uncertainty} & 81.1  & 11.2 &  80.1& 13.7& 87.1 & 18.1 & 77.6   & 14.7\\
    &ICT \cite{verma2022interpolation} & 82.4 & 7.2 & 81.5  & 7.8  & 87.6 & 10.6 & 78.2 & 3.2\\
    &CCT  \cite{ouali2020semi} & 84.0 & 6.6 & 82.3  & 5.4 & 88.6& 
9.4 & 81.0 & 5.1 \\
    &CPS \cite{Chen2021a} & 85.0& 6.6& 82.9   &6.6 & 88.0 &10.8 & 84.2 & 2.3\\
    &CTS  \cite{Luo2022} & 86.4 & 8.6 & 84.4 & 6.9  & 90.1 & 11.2 & 84.8 & 7.8\\
     & MCSC  \cite{Liu_2023_BMVC}  & 89.4 &  2.3 &  {\bf87.6} & {\bf1.1}   & {\bf93.6} &  3.5 &  87.1 &  2.1\\
    & Ours (CNN, Affine) & \underline{89.5} & \underline{1.8} &  87.2 &  2.0   & \underline{92.9} &  \bf{1.8} &  88.4 &  \underline{1.7}\\
    & Ours (Trans, Affine) & 89.1  &  \underline{1.8} &   85.7 & \underline{1.2}   & 91.7 &  2.8 & \underline{89.9} &  \bf{1.3}\\
    & Ours (mean, Affine) & \bf{90.3} & \bf{1.6} & \underline{87.4} & 1.4 & 92.7 & \underline{2.2} & \bf{90.9} & \bf{1.3} \\
     \midrule
     \multirow{14}{*}{ 3 (5$\%$)}&Reg. only (Aff) & 32.0 & 17.8 & 18.0 & 15.7 & 43.9 & 16.0 & 34.0 & 21.7 \\
    \cmidrule(r){2-10} 
     & DeepAtlas \cite{xu2019deepatlas}&  59.0 & 8.6 & 62.8& 5.4 & 67.8 & 7.7 & 46.4 & 12.6 \\
      \cmidrule(r){2-10} 
     & UNet-LS  & 51.2   & 31.2 & 54.8  & 24.4 & 61.8  &24.3& 37.0 & 44.4  \\
     &MT \cite{tarvainen2017mean}  & 56.6 &34.5& 58.6 &23.1&	70.9&26.3 &40.3&53.9\\
    &DCT \cite{qiao2018deep}  & 58.2 & 26.4 & 61.7 & 20.3 & 71.7& 27.3 & 41.3 & 31.7\\
    &UAMT  \cite{yu2019uncertainty}  & 61.0  & 25.8 & 61.5   & 19.3&  70.7& 22.6& 50.8 & 35.4\\
    &ICT \cite{verma2022interpolation} & 58.1 & 22.8 & 62.0  & 20.4  & 67.3 & 24.1 & 44.8 & 23.8\\
    &CCT  \cite{ouali2020semi}  & 58.6 & 27.9 & 64.7  & 22.4& 70.4& 27.1 & 40.8 &34.2 \\
    &CPS \cite{Chen2021a}   & 60.3& 25.5& 65.2   &18.3 & 	72.0 &22.2 & 43.8 & 35.8 \\
    &CTS \cite{Luo2022}   & 65.6 & 16.2 & 62.8 &11.5  & 76.3 & 15.7 & 57.7 & 21.4\\
    & MCSC  \cite{Liu_2023_BMVC} & 73.6  & 10.5  & 70.0 & 8.8 & 79.2& 14.9& 71.7 & 7.8\\
    & Ours (CNN, Affine) & 85.2  & \bf{1.9}  & \underline{83.3} & \underline{1.5} & \bf{89.9}&  \underline{2.9} & \underline{82.4} & \underline{2.2}\\
    & Ours (Trans, Affine) & \underline{85.4}  & 2.6  & 83.2 & 1.8 & \underline{89.3}& 3.8 &\bf{83.5} &\bf{2.1}\\
    & Ours (mean, Affine) & \bf{85.7} & \underline{2.0} & \bf{83.8} & \bf{1.4} & \bf{89.9} & \bf{2.4} & \bf{83.5} & \bf{2.1} \\

    \midrule
    \multirow{8}{*}{ 1 (1.4\%) } &Reg. only (Aff) & 23.4 & 19.7 & 13.6 & 18.7 & 31.6 & 19.0 & 25.1 & 21.4 \\ 
    \cmidrule(r){2-10} 
     & DeepAtlas \cite{xu2019deepatlas}& 40.4 & 18.5 & 42.2 & 11.7 & 34.7 & 29.2 & 44.4 & 14.6    \\
      \cmidrule(r){2-10} 
    & UNet-LS  &  26.4   & 60.1 & 26.3  & 51.2 & 28.3  & 52.0& 24.6 & 77.0  \\

    &CTS  \cite{Luo2022}  & 46.8 &  36.3  & 55.1 & 5.5&	64.8 & {\bf4.1} &20.5 & 99.4\\
     &MCSC  \cite{Liu_2023_BMVC} & 58.6  & 31.2& 64.2 &13.3&	78.1 &  12.2 & 33.5 & 68.1\\
       &Ours (CNN, Affine)  & 79.6   & 5.2 &77.6 & 5.3&	\underline{83.2} & 5.1 & 78.0 &5.1\\
       &Ours (Trans, Affine) & \underline{80.0} & \underline{4.2} & \underline{77.7} & \underline{4.0}&	83.0 &\underline{4.2} & \bf{79.4} & \underline{3.6}\\
       & Ours (mean, Affine) & \bf{80.4} & \bf{3.5} & \bf{78.3} & \bf{3.2} & \bf{83.6} & 4.3 & \underline{79.3} & \bf{2.9} \\
    \bottomrule 
    \multicolumn{10}{r}{\scriptsize \textbf{Best} is bold, \underline{Second Best} is underlined. \hfill}
    \end{tabular}
    }
    \label{tab:full_acdc}
\end{table*}

\begin{table*}
	\caption{Segmentation results on Synapse for our method CCT-R and baselines, according to DSC($\%$) and HD(mm).}
	% The performance is reported by DSC (\%) and HD (\%), as well as the DSC value of each types of organs.
	\renewcommand*{\arraystretch}{1}
	\setlength{\tabcolsep}{2pt}
	\centering
	% \resizebox{0.75\linewidth}{!}
 {
 \small
		\begin{tabular}{@{}c c c c c c c c c c c c@{}}
			\toprule
			\textbf{Labeled} & \textbf{Methods} & DSC$\uparrow$ & HD$\downarrow$ & Aorta & Gallb & Kid$\_$L & Kid$\_$R & Liver & Pancr & Spleen & Stom \\
			\midrule
			\multirow{2}{*}{18(100$\%$)} &  UNet-FS  & 75.6 & 42.3 & 88.8 & 56.1 & 78.9 & 72.6 & 91.9 & 55.8 & 85.8 & 74.7 \\
			&  nnFormer & 86.6 & 10.6 & 92.0 & 70.2 & 86.6 & 86.3 & 96.8 & 83.4 & 90.5 & 86.8 \\
			\midrule
			\multirow{16}{*}{4(20$\%$)} & Reg. only (Affine) & 27.0 & 39.6 & 16.0 & 7.5 & 36.4 & 33.0 & 56.8 & 13.1 & 28.5 & 25.1  \\
                & Reg. only (Aff+Def) &   32.5  & 36.5  & 29.7  & 4.8  & 36.5 & 29.4  & 65.5 & 14.2  & 48.0  & 31.7   \\
                \cmidrule(r){2-12}
                  & DeepAtlas \cite{xu2019deepatlas}& 56.1&85.3&69.2&43.3&50.8&55.2&88.8&30.5&62.7&48.0 \\
                  \cmidrule(r){2-12}

			& UNet-LS & 47.2 & 122.3 & 67.6 & 29.7 & 47.2 & 50.7 & 79.1 & 25.2 & 56.8 & 21.5 \\
			& UAMT \cite{yu2019uncertainty}  & 51.9 & 69.3 & 75.3 & 33.4 & 55.3 & 40.8 & 82.6 & 27.5 & 55.9 & 44.7 \\
			& ICT \cite{verma2022interpolation} & 57.5 & 79.3 & 74.2 & 36.6 & 58.3 & 51.7 & 86.7 & 34.7 & 66.2 & 51.6 \\
			& CCT \cite{ouali2020semi} & 51.4 & 102.9 & 71.8 & 31.2 & 52.0 & 50.1 & 83.0 & 32.5 & 65.5 & 25.2 \\
			& CPS \cite{Chen2021a} & 57.9 & 62.6 & 75.6 & 41.4 & 60.1 & 53.0 & 88.2 & 26.2 & 69.6 & 48.9 \\
			% & CTS(CNN) & 62.8 & 91.2 & 79.1 & 30.3 & 69.2 & 66.3 & 84.8 & 42.4 & 76.5 & 53.5 \\
			&CTS\cite{Luo2022} & 64.0 &  56.4 & 79.9 & 38.9 & 66.3 & 63.5 & 86.1 & 41.9 & 75.3 &60.4 \\
                & MCSC \cite{Liu_2023_BMVC}& 68.5 & 24.8 & 76.3 & \underline{44.4} & \underline{73.4} &  \underline{72.3} & 91.8 &46.9 & 79.9 & 62.9 \\
                & Ours (CNN, Affine)&67.3 & 37.9 & 79.0 & 36.5&72.7 &70.4	& 87.9 &	47.3 & 77.8 &	67.0 \\
                & Ours (Trans, Affine)&70.5 &22.7 & \bf{81.0} & 34.1 &	71.1 & 71.9 & 93.2 & \underline{49.9} & \bf{87.9} & \bf{75.2} \\
                & Ours (mean, Affine)&70.0& 23.2& 79.8 &34.5& 71.0 & 70.7 &92.8 & 49.6 &\underline{87.4} & \underline{74.4} \\
			& Ours (CNN, Affine+Deform)  & 69.5 & 36.2 & 80.0 & \bf{49.2} & 73.0 & 69.9 & 89.3 & 48.5 & 79.5 & 66.7 \\
           & Ours (Trans, Affine+Deform)  & \bf{72.5} & \bf{20.5} & \underline{80.9} & 43.4 & \bf{75.6} & \bf{75.1} & \underline{93.5} & \bf{51.3} & \underline{87.4} & 72.2 \\
		   & Ours (mean, Affine+Deform)  & \underline{71.4} & \underline{21.1} & 80.4 & 42.3 & 73.0 & 70.0 & \bf{93.7} & 49.4 & \bf{87.9} & 74.2 \\
			\midrule
			\multirow{16}{*}{2(10$\%$)} & Reg. only (Affine) &25.4&36.8&17.5 & 3.5 & 32.7 & 27.5 &53.4 & 12.6 & 33.4 &22.5  \\
                & Reg. only (Aff+Def) &  29.1 & 44.0 & 27.2 & 11.3 & 28.6 & 26.5 & 66.4 & 12.7 & 29.7 & 30.3    \\
                \cmidrule(r){2-12}
                & DeepAtlas \cite{xu2019deepatlas}& 44.0& 67.1& 68.0  & 24.9 & 37.9  & 46.0 & 82.7  & 18.4 & 44.2  &30.6  \\
                  \cmidrule(r){2-12}
			& UNet-LS & 45.2 & 55.6 & 66.4 & 27.2 & 46.0 & 48.0 & 82.6 & 18.2 & 39.9 & 33.4 \\
			& UAMT \cite{yu2019uncertainty} & 49.5 & 62.6 & 71.3 & 21.1 & 62.6 & 51.4 & 79.3 & 22.8 & 58.2 & 29.0 \\
			& ICT \cite{verma2022interpolation} & 49.0 & 59.9 & 68.9 & 19.9 & 52.5 & 52.2 & 83.7 & 25.4 & 53.2 & 36.0 \\
			& CCT \cite{ouali2020semi} & 46.9 & 58.2 & 66.0 & 26.6 & 53.4 & 41.0 & 82.9 & 21.2 & 48.7 & 35.6 \\
			& CPS \cite{Chen2021a} & 48.8 & 65.6 & 70.9 & 21.3 & 58.0 & 45.1 & 80.7 & 23.5 & 58.0 & 32.7 \\
			% & CTS(CNN) & 51.3 & 64.4 & 75.3 & 17.3 & 58.8 & 65.4 & 78.8 & 20.8 & 71.4 & 22.5 \\
			& CTS \cite{Luo2022}  & 55.2 & 45.4 & 71.5 & 25.6 & 62.6 & 67.5 & 78.2 & 26.3 & 75.9 & 34.3 \\
                & MCSC \cite{Liu_2023_BMVC}&  61.1 &  32.6 & 73.9 & 26.4 &  69.9 & 72.7 & 90.0 & 33.2 & 79.4 & 43.0  \\
                & Ours (CNN, Affine)& 60.4	& 37.1&	77.0	&27.8&70.8&69.0	&88.4&35.4&	67.0	&47.7 \\
               & Ours (Trans, Affine)& 64.2	&\underline{22.1}&\underline{77.4}&22.1&75.0&74.2&92.2&\underline{39.6}&78.2&54.8 \\
               & Ours (mean, Affine)& 65.1	&22.5&75.7&	28.4	&74.5&\bf{75.0}&91.8&38.0&\underline{82.3}&55.1 \\
			& Ours (CNN, Affine+Deform)  & 62.6 & 44.3 & 76.5 & \underline{37.7} & 73.0 & 68.0 & 87.0 & 32.3 & 76.5 & 49.9 \\
           & Ours (Trans, Affine+Deform)  & \bf{68.3} & 23.1 & 74.8 & \bf{49.1}  & \bf{75.2} & \underline{74.7} & \bf{92.8} & \bf{39.7} & \bf{84.1} & \bf{56.2} \\
			& Ours (mean, Affine+Deform)  & \underline{66.5} & \bf{19.7} & \bf{77.6} & 34.4 & \underline{75.1} & 74.2 & \underline{92.6} & 39.5 & 82.1 & \underline{56.1} \\
			\midrule
			\multirow{16}{*}{1(5$\%$)} & Reg. only (Affine) & 26.4  & 45.0  & 16.3  & 6.6  & 35.8 & 32.8  & 53.5 & 14.4  & 28.7  & 22.7  \\
                & Reg. only (Aff+Def) &  27.4  & 52.2  & 26.4  & 11.3  & 30.5 & 27.1  & 61.6 & 12.8  & 26.3  & 23.6   \\
                \cmidrule(r){2-12}
            & DeepAtlas \cite{xu2019deepatlas}&  16.1 & 72.3 & 18.4 & 14.9  &1.2  & 10.1  & 57.1 & 0.6 &14.4 &12.2    \\
                  \cmidrule(r){2-12}
			& UNet-LS & 13.7 & 116.5 & 11.6 & \bf{17.8} & 0.8 & 1.8 & 56.9 & 0.1 & 8.7 & 11.6 \\
			& UAMT\cite{yu2019uncertainty}  & 10.7 & 90.2 & 8.0 & 9.3 & 0.3 & 8.1 & 31.7 & 1.1 & 13.1 & 14.3 \\
			& ICT\cite{verma2022interpolation} & 15.9 & 82.3 & 13.8 & 11.9 & 0.3 & 2.7 & 70.5 & 0.8 & 16.4 & 10.6 \\
			& CCT\cite{ouali2020semi} & 11.7 & 107.5 & 10.0 & 13.0 & 0.1 & 1.9 & 47.5 & 3.7 & 8.0 & 9.3 \\
			& CPS \cite{Chen2021a} & 15.0 & 123.5 & 19.6 & 9.6 & 5.6 & 6.9 & 59.4 & 2.3 & 9.4 & 7.2 \\
			% & CTS(CNN) & 25.1 & 95.4 & 47.8 & 0 & 3.1 & 3.1 & 64.5 & 10.5 & 49.0 & 22.7 \\
			& CTS \cite{Luo2022}& 26.3 & 96.5 & 44.6 & 4.0 & 11.2 & 5.5 & 60.3 & 9.6 & 54.1 & 21.2 \\
                & MCSC \cite{Liu_2023_BMVC}&34.0 & 53.8 & 50.9 & 13.0 & 17.6& 54.6 & 64.3 & 5.5 & 43.1 & 23.5 \\ 
			& Ours (CNN, Affine)  & 39.5  & 66.5 & 61.7 & \underline{17.0} & 9.2 & 65.2 & 71.1 & \bf{12.3} & 54.3 &  25.3\\
           & Ours (Trans, Affine)  &43.2  &67.5  & 58.5 &  12.5  & 20.2&  66.6  & \bf{78.9} &  10.3  & \underline{72.9}  & 26.5  \\
			& Ours (mean, Affine)  & 43.4 & \underline{40.8} & 62.5 & 13.3 & 17.9 & \bf{71.0} & \underline{77.0} & \underline{11.4} & 65.4 & \bf{28.7} \\
            & Ours (CNN, Affine+Deform)& 44.2&54.2&\underline{63.8}&10.8&48.7&61.6&74.6&5.4&	61.8&26.6 \\
            & Ours (Trans, Affine+Deform)& \underline{45.3}	&46.9&62.9&	9.9&\underline{56.5}&65.6&70.9&	0.1&72.8	&24.2 \\
            & Ours (mean, Affine+Deform)& \bf{47.6} &\bf{38.4}&\bf{65.5}&9.3&\bf{61.6}&\underline{70.2}&72.7&	0.1&\bf{73.9}	&\underline{27.8} \\
			\bottomrule
			\multicolumn{12}{r}{\scriptsize \textbf{Best} is bold, \underline{Second Best} is underlined. \hfill}
		\end{tabular}
	}
	\label{tab:full_synapse}
\end{table*}

\end{document}